\documentclass[letterpaper, 10 pt, journal, twoside]{IEEEtran}

\usepackage{multicol}
\usepackage[bookmarks=true]{hyperref}
\usepackage[noadjust]{cite}
\usepackage{booktabs}
\usepackage{graphicx}
\graphicspath{{./images/}}
\usepackage{float}
\usepackage{amsmath, amssymb}
\usepackage{amsfonts}
\usepackage{bm}
\usepackage{algpseudocode}
\usepackage{algorithm}
\usepackage{multirow}

\usepackage[skip=1pt,font=small,labelfont=bf]{caption}
\usepackage{subcaption}
\usepackage{rotating}
\usepackage{stfloats}

\usepackage{adjustbox}
\usepackage{tikz}
\usepackage{pgfplots}

\definecolor{lightgreen}{RGB}{240,249,232}
\definecolor{darkblue}{RGB}{43,140,190}

\hypersetup{
    colorlinks=true,
    linkcolor=black,
    urlcolor=blue,
}
\newcommand{\rebuttal}[1]{{#1}}

\newcommand{\ov}{\bm{o}}

\newcommand{\pv}{\bm{p}}
\newcommand{\uv}{\bm{u}}
\newcommand{\qv}{\bm{q}}

\newcommand{\sdf}{\texttt{SDF}}
\newcommand{\sdfhat}{\widehat{\sdf}}
\newcommand{\selfcoll}{\texttt{self\_coll}}

\newcommand\numberthis{\addtocounter{equation}{1}\tag{\theequation}}
\newcommand{\shrinka}{\def\baselinestretch{0.993}\large\normalsize}

\usepackage{amsmath}
\DeclareMathOperator*{\argmax}{arg\,max}

\usepackage{relsize}
\usepackage{todonotes}

\definecolor{international_orange}{RGB}{240, 74, 0}
\definecolor{pacific_blue}{RGB}{2,108,181}

\begin{document}
\title{Planning Visual-Tactile Precision Grasps via Complementary Use of Vision and Touch} 
\markboth{IEEE Robotics and Automation Letters. Preprint Version. Accepted December, 2022}{Matak \MakeLowercase{\textit{et al.}}: Planning Visual-Tactile Precision Grasps via Complementary Use of Vision and Touch}

\author{Martin Matak$^{1}$ and Tucker Hermans$^{1,2}$ 
\thanks{Manuscript received: September 2, 2022; Revised November 18, 2022; Accepted December 14, 2022.}
\thanks{This paper was recommended for publication by Hong Liu upon evaluation of the Associate Editor and Reviewers' comments.
This work was supported by supported by NSF Award \#1846341, by DARPA grant N66001-19-2-4035, and by a Sloan Research Fellowship.}
\thanks{$^{1}$ Martin Matak and Tucker Hermans are with the School of Computing and the Robotics Center, University of Utah, Salt Lake City, UT, USA. {\tt\footnotesize martin.matak@utah.edu} $^{2}$ NVIDIA; Seattle, WA, USA}
\thanks{Digital Object Identifier (DOI): see top of this page.}
}

\everypar{\looseness=-1}

\input{intro-fig}
\maketitle

\begin{abstract}
  Reliably planning fingertip grasps for multi-fingered hands lies as a key challenge for many tasks including tool use, insertion, and dexterous in-hand manipulation.
  This task becomes even more difficult when the robot lacks an accurate model of the object to be grasped.
  Tactile sensing offers a promising approach to account for uncertainties in object shape.
  However, current robotic hands tend to lack full tactile coverage.
  As such, a problem arises of how to plan and execute grasps for multi-fingered hands such that contact is made with the area covered by the tactile sensors. To address this issue, we propose an approach to grasp planning that explicitly reasons about where the fingertips should contact the estimated object surface while maximizing the probability of grasp success.
Key to our method's success is the use of visual surface estimation for initial planning to encode the contact constraint.
  The robot then executes this plan using a tactile-feedback controller that enables the robot to adapt to online estimates of the object's surface to correct for errors in the initial plan. \rebuttal{Importantly, the robot never explicitly integrates object pose or surface estimates between visual and tactile sensing, instead it uses the two modalities in complementary ways. Vision guides the robots motion prior to contact; touch updates the plan when contact occurs differently than predicted from vision.}
  We show that our method successfully synthesises and executes precision grasps for previously unseen objects using surface estimates from a single camera view. Further, our approach outperforms a state of the art multi-fingered grasp planner, while also beating several baselines we propose.
\end{abstract}

\begin{IEEEkeywords}
Multifingered Hands; Grasping; Deep Learning in Grasping and Manipulation
\end{IEEEkeywords}

\section{Introduction}
\label{sec:intro}
\IEEEPARstart{I}{n} this letter we look into enabling a robot to pick up everyday objects that the robot hasn't encountered before. We focus on the task of grasping previously unseen objects as a fundamental step toward successfully deploying robots in household environments. In such environments, the robot should be able not to only pick up objects, but also to use them for everyday tasks~\cite{sundaralingam-ral2020-benchmarking-in-hand}. This requires the ability of fine in-hand manipulation of objects, which makes multi-fingered hands a good candidate as a general end effector. While multi-fingered hands seem promising for downstream tasks, using them to grasp everyday objects and ensuring the resulting grasp makes contact only with the fingertips remains a challenging problem~\cite{lu-ral2019-grasp-type, lu-iros2020-active-grasp}.

It's obvious that contact with an object must be made in order to successfully pick the object up, but it's unclear how to explicitly enforce this constraint when a robot doesn't know the complete object shape. This is the core focus of this paper---establishing an in contact, precision grasp with a previously unseen object when given a single RGBD image from a fixed camera-view. To help account for the lack of information a single camera view provides, we examine how tactile sensing can improve the grasp control process. In particular we examine how to do this when the robot only has tactile sensing on its distal links (fingertips), a common setup on contemporary hands~\cite{yamaguchi-humanoids2016-fingervision,hogan-iros2018-tactile-regrasp,bala-icra2019-tactile-force-estimation}.

In general, tactile sensors improve grasping. Calandra and colleagues~\cite{calandra-corl2017-feeling-of-success} show that tactile feedback, in addition to vision-based input, provides more information for computing a good grasp, while Murali et al.~\cite{murali-iser2018-without-seeing} present an approach to grasping using tactile sensing without vision. However, it's unclear how to get to a grasp such that tactile feedback is registered, i.e. the contact is made with the tactile sensors when planning with fully actuated, dexterous hands. Having only limited tactile coverage on most robotic hands doesn't help either. One approach~\cite{lu-ral2019-grasp-type} is to explicitly model a \textit{precision} grasp, which would result in a grasp where contact with the object is made with fingertips of the hand, exactly where the tactile sensors are. After the hand is in contact with the object, Veiga et al.~\cite{veiga-sensors2020-finger-control} show how to stabilize the grasp using tactile feedback. Apart from the tactile-guided approaches, there are many vision-based methods to tackle grasping~\cite{bohg-tro2014-grasp-survey}. Several of these use 3D reconstruction from partial views to aid the grasp planning process~\cite{vandermerwe-icra2020-reconstruction-grasping,lundell-2021icra-fingan,jiang2021synergies} and others examine reconstructing object shape from touch~\cite{dragiev-icra2013-uncertainty-grasping-tactile,murali-iser2018-without-seeing,yi-iros2016-active-touch}. However, no work has shown the ability to plan, adapt, and ensure precision grasps for multi-fingered hands through joint visual and tactile sensing \rebuttal{in the real world.} Indeed, as discussed in Section~\ref{sec:related-work}, most multi-fingered grasping approaches use simple, heuristic controllers to move from a grasp preshape to an in-contact grasp. \rebuttal{Further, similar works evaluated in simulation make unrealistic assumptions such as full tactile coverage of the robot and the ability to repeatedly make contact with the object without it moving}.

\rebuttal{To address this gap, we propose a learning-guided and geometry informed grasp controller to autonomously generate precision grasps on previously unseen objects.}
We formulate grasp synthesis as an optimization problem with a hard constraint that contact must be made between the estimated object surface and the robot's tactile fingertips.
The robot uses a neural network that encodes the probability of grasp success as its objective function as in~\cite{lu-ram2020-MultiFingeredGP}.
We generate an initial estimate of the object surface from the partial-view point cloud using the PointSDF~\cite{vandermerwe-icra2020-reconstruction-grasping} neural network.

While the estimated object surface and pose provide a model useful for planning a grasp, we also know it is very likely inaccurate. To account for this we leverage tactile feedback during the grasp control phase to adapt to the uncertain object shape and ensure contact. Notably, we do not fuse the visual and tactile surface estimates into a single model, instead we use them to separately plan and ensure contact. As such we avoid the difficult problem of registering the different measurements under extrinsic calibration error and possible object motion. After closing the hand,
the robot evaluates an in-contact grasp classifier to decide whether to lift the object. Figure~1 visualizes all stages of our approach.

\rebuttal{
  We experimentally demonstrate that our approach can reliably generate multi-fingered precision grasps for novel objects without explicitly integrating vision and touch. We run real world robot experiments that quantify that using both vision and touch improves grasp performance over using vision alone. Our results additionally characterize the benefits of our approach over prior work. We successfully plan and execute precision grasps for various, novel objects observed from a single camera view. Some examples of our grasps are shown in Fig.~\ref{fig:resulting-grasps}. Experiments videos and source code can be found on our project website: \url{https://sites.google.com/view/precision-grasps}.}

\section{Related work}
\label{sec:related-work}
We divide our discussion of previous grasp planning work between geometry-based and learning-based approaches before discussing grasping with tactile sensing.

\subsection{Geometry-based Grasp Synthesis}
Geometric-based grasping approaches leverage a model of the object geometry in order to compute a grasp metric.
One of the most popular approaches in this category is Eigengrasp \cite{ciocarlie-ijrr2009-eigengrasp}. Eigengrasp computes a grasp pose and joint configuration in a low-dimensional ``eigengrasp'' space and closes the fingers at equal joint velocity rates until detecting contact. Assuming full knowledge of object shape, Roa and Suarez~\cite{suarez-2009tro-contact-regions} determine contact regions on the object to make contact with, such that the resulting grasp is in force closure. Rosales et al.~\cite{suarez-2011ijrr-synthesizing-grasps} solve for hand configuration to be in contact with pre-specified regions on the object. Zheng and Qian~\cite{zheng-2005ijrr-force-closure-uncertainty} quantify friction and contact position uncertainty when computing force closure grasps.

Another geometry-based, uncertainty aware grasp synthesis approach comes from \cite{chen-2018-pSDF}, which fuses multiple camera view to infer a ``pre-touch'' configuration to maximize grasp success. Hang et al.~\cite{hang-tro2016-hierarchical-fingertip-space} synthesize a grasp that is in contact with an object, and once the hand is in contact, use a learned probabilistic controller to stabilize it. Siddiqui~\cite{siddiqui-frontiers2021-bayesian-exploration} use a Bayesian inspired method to efficiently explore a variety of grasps. After moving to a predicted pose, the hand is closed using a simple controller until contact is detected. Morales et al.~\cite{morales-iros2006-asfour} build a database of possible power and precision grasps and retrieve a grasp for a target object at run time. They obtain precision grasps using a simple heuristic controller, but execute no grasps in the real world. Planning to grasp an object with uncertain pose, but known geometry, has also been examined formally as a belief-space planning problem~\cite{platt2011simultaneous}, however, scaling this to be efficient with uncertainty in shape, while likely useful, is an open challenge. In the case of uncertain shape, such as when estimated from a single camera view as is our focus, these geometry-based grasp metrics will likely be inaccurate or suboptimal due to inaccuracies in estimated shape.  
\vspace{-8pt}
\subsection{Learning-based Grasp Synthesis}
The research literature contains numerous data-driven grasp approaches~\cite{kappler-icra2015-leveraging-big-data, varley-iros2015-generating-multi-fingered, saxena-aaai2008-learning-grasp, kopicki-ijrr2019-better-generative-models, calandra-corl2017-feeling-of-success, merzic-icra2019-leveraging-contact-forces, lu-iros2020-active-grasp}. We build on the approach from~\cite{lu-ram2020-MultiFingeredGP} to generate a grasp preshape. The primary novelty of our work comes after the robot moves to this preshape.
We classify learning-based grasp synthesis approaches between those that generate grasps in contact with the object and grasps that are close to, but not in contact with the object. We found no work that combines the two.

\textbf{In contact grasps:}
Varley et al.~\cite{varley-iros2015-generating-multi-fingered} train a CNN to predict heatmaps for fingertip locations on the object as well as palm pose, which are used to synthesize a grasp.
While Veres et al.~\cite{veres-ral17-grasp-motor-imagery} generate contact points on an object for a 3-fingered hand using generative models (in simulation), Shao et al.~\cite{shao-ral2020-unigrasp} manage to synthesize grasps as contact points on an object for various grippers and demonstrate their approach in a real-world setting.
Sundermeyer and colleagues~\cite{sundermeyer-icra2021-contact-graspnet} predict grasp locations such that one finger of a parallel jaw gripper makes contact with the observed point cloud, this allows for a clever reparameterization of the 6DOF prediction to 4DOF. However the second finger location is determined based on a predicted gripper width, so it is unclear how to scale this approach to fully actuated multi-fingered hands. All approaches in this category predict grasps in contact with the object and are therefore brittle to shape estimation and camera noise. To resolve this, other approaches predict grasps near the object, but not in contact. 

\textbf{Near contact grasps:}
Kappler et al.~\cite{kappler-icra2015-leveraging-big-data} predict grasp as a palm pose,
\rebuttal{Wu} et al.~\cite{wu-iros2019-pixel-attentive} use a policy gradient method to predict grasp pose and joint angles from a depth image. Lundell and colleagues~\cite{lundell-2021ral-clutter, lundell-2021icra-fingan} use a GAN to predict grasps that are almost on the surface of the object, yet not in contact. Similar to~\cite{lu-iros2020-active-grasp}, Mousavian et al.~\cite{mousavian-iccv2019-graspnet} initialize and refine a grasp pose using a learned neural net. \rebuttal{Our own previous work~\cite{lu-ral2019-grasp-type, lu-ram2020-MultiFingeredGP,lu-iros2020-active-grasp,vandermerwe-icra2020-reconstruction-grasping} synthesizes a preshape similarly as in this letter, but doesn't explicitly reason about making contact either through vision or touch}. All the approaches in this category after moving the robot to the preshape use either a simple heuristic controller to close the hand or don't specify how exactly the hand is closed. We identify this as a gap in the literature and explore it here. For a deeper overview of dexterous grasping, we refer the reader to~\cite{duan-frontiers2021-dexterous-grasping-summary}.

\subsection{Tactile Enabled Grasp Evaluation and Control}
Calandra and colleagues~\cite{calandra-ral2018-more-than-feeling} propose a grasping approach that maximizes grasp success probability using a sampling based method. Their model predicts success probability based on tactile 
readings, RGB image, and a regrasping action for a two-fingered gripper. If the predicted grasp success probability is above a threshold (90\%), the robot attempts to lift the \rebuttal{object}. 
Hogan et al.~\cite{hogan-iros2018-tactile-regrasp} follow a similar approach but use a tactile-only grasp quality metric. For their resampling strategy they simulate various actions (movements of the gripper) and pick the action with the highest probability of success, as predicted by the tactile-only metric. Wu et al.~\cite{wu-corl2019-mat} use a deep reinforcement learning method to implement a closed-loop approach to grasping. They use tactile sensors on the 3-fingered Barett Hand as part of the observations used by the RL policy. Dang and Allen~\cite{dang-ar2014-stable-grasping} regrasp by searching for the closest stable grasp in the database of all the previous grasps performed by the robot. After finding such a grasp, they adjust the hand. Similar to the rest of the work in this section, the tactile readings are used for the grasp quality metric, but also to find similar grasps in the database. Murali et al.~\cite{murali-iser2018-without-seeing} use learned corrective actions that map tactile input in latent space to gripper adjustment in task space. Dragiev et al.~\cite{dragiev-icra2013-uncertainty-grasping-tactile} develop a tactile-informed regrasp control law. \rebuttal{Farias et al.~\cite{farias-ral21-gpis-tactile-vision} estimate object shape from vision, update the shape estimate using tactile feedback, and use the resulting model to estimate force closure. Both papers~\cite{farias-ral21-gpis-tactile-vision,dragiev-icra2013-uncertainty-grasping-tactile} only evaluate in simulation and assume the robot can detect contact anywhere on the hand.}

\section{Problem Description}
\label{sec:problem}
The problem we wish to solve has two main phases. First, the robot plans a precision grasp that maximizes the probability of grasp success under uncertain object surface estimation. Then, the robot must execute the planned grasp using tactile feedback to compensate for shape uncertainty.

Let $f(\qv, \ov)$ estimate the probability of grasp success for configuration $\qv$ for object $\ov$. Let the robot have $n$ degrees of freedom (DoF) in the arm and $m$ DoF in the hand. We are searching for the robot configuration $\qv \in \mathbb{R}^{n+m}$ such that every fingertip is on the surface of the object $\ov$ and that $f$ is maximized. The surface of the object $\ov$ is estimated from a partial pointcloud. Let $\phi_p^i:\mathbb{R}^4 \rightarrow \mathbb{R}^3$ denote the forward kinematics function for the position of the central point on the surface of the fingertip $i$. Input to $\phi_p^i$ is in $\mathbb{R}^4$ because every finger on the hand we work with has 4 DoF. 
Let $\sdf(p,\ov)$ be the signed distance function between a point $p \in \mathbb{R}^3$ and the object $\ov$\rebuttal{, returning positive values for \(p\) outside the object, negative inside, and 0 on the surface.}

We formalize the search for the grasp configuration $\qv^*$ as
\begin{align*}
\qv^* =   \argmax_{\qv} \quad & f(\qv, \ov) \numberthis \label{eq:objective-function} \\
  \text{s.t.}   \quad & \qv_{min}\preceq \qv \preceq \qv_{max} \numberthis \label{eq:jointlimits-constraint} \\
  		\quad & \selfcoll(\qv) = 0 \numberthis \label{eq:self-collision-constraint} \\
  		\quad & \sdf(\phi_p^i(\qv), \ov) = 0 \; \forall i \in \{1,\ldots,F\} \numberthis \label{eq:sdf-constraint}.
\end{align*}
Equation~(\ref{eq:jointlimits-constraint}) constrains the solution within joint limits, while the function $\selfcoll(\cdot)$ counts the number of hand links in self-collision and $F$ denotes the number of fingers we wish to make contact with the object. In our experiments $F=4$. 

A solution to the optimization problem defines a grasp such that fingertips are in contact with the object surface and not in self collision. However, the estimated object shape is likely imperfect and hence there is inherently a mismatch between estimated object surface and the real one. This defines the second part of the problem - moving the robot to the planned grasp knowing a contact with the object might happen before or after expected. Not accounting for this may result in knocking the object over or stopping before contact is made. Formally, given current configuration $\qv[k]$, tactile readings \(\tau[k]\), and target configuration $\qv_F$, the robot must execute a control policy $\uv = \pi(\qv[k], \tau[k], \qv_F)$ to achieve a grasp where its tactile fingertips contact the object.

\section{Our Approach}
\label{sec:approach}
Our approach consists of four steps: (1) planning and moving the robot arm and hand to a grasp preshape configuration, (2) planning an in-contact grasp using the estimated object shape, (3) moving the robot fingers to the planned in-contact grasp configuration and (4) evaluating the robot's in-contact classifier to decide whether to lift the object or regrasp it. The first two steps base decisions purely on point cloud input, whereas the third step uses tactile feedback to adapt the plan online. The fourth step uses visual input and the current robot configuration. We do not fuse visual and tactile sensing to make a unified model of the object surface, instead we use the two separately to satisfy the contact constraint during planning and control respectively. Figure~1 visualizes these steps. We now describe the four components in turn, followed by implementation details of our learned models.

\subsection{Grasp Preshape Synthesis}
As the first step in our approach we plan a grasp preshape $\qv_0 = [\qv_A, \qv_H]$ which maximizes the probability of grasp success, but does not make contact with the object. The preshape contains both the arm configuration $\qv_A \in \mathbb{R}^n$ and the hand configuration $\qv_H \in \mathbb{R}^{m}$. We solve for the preshape using the approach from~\cite{lu-ram2020-MultiFingeredGP} which maximizes the probability of grasp success using a learned neural network as its objective.
The objective contains two terms: a mixture density network (MDN) as a prior, and a preshape classifier as a likelihood component. The MDN additionally serves as a sampler to initialize the optimization. Both neural networks take as input a voxelized representation of the object from the observed single-view RGBD image. We give further details about the networks and their training below.
The preshape planner serves to move the palm of the hand to a pose close to the object and the proximal joints of each finger into a configuration amenable to making a high quality precision grasp. As in~\cite{lu-ram2020-MultiFingeredGP} we optimize using bound constrained BFGS\rebuttal{~\cite{bfgs}}.
Figure 1 left shows an example grasp preshape.

\subsection{In-Contact Grasp Synthesis and Evaluation}
After moving to the preshape configuration $\qv_0$, the robot plans to make contact with the object. The robot uses \(\qv_0\) to initialize its optimization for a hand configuration in contact with the object. This differs from the previous step, where we optimize for arm and hand configuration. For the rest of the approach, the arm configuration $\qv_A$ remains fixed and we only change the joints of the fingers $\qv_H$.

We reformulate the in-contact grasp synthesis problem from Eqs.~(\ref{eq:objective-function})-(\ref{eq:sdf-constraint}), by relaxing the SDF constraint and using an estimate of the SDF, \(\sdfhat{}\). We add an additional term to the objective to encourage alignment between the fingertip surface normal and the approach vector resulting in the following penalty-method objective, \(\tilde{f}(\qv, \ov, \pv)=\)\vspace{-6pt}
\begin{equation}
  \sum_{i=1}^F||\mathsmaller{\sdfhat{}(\pv_i, \ov)}||^2 + ||\phi_{p}^{i}(\qv) - \pv_{i}||^2 + \alpha ||1-\cos\theta_{i}||^2\label{eq:impl-obj-f}\vspace{-6pt}
\end{equation}
where $\theta_{i}$ is angle \rebuttal{between the normal on \(i\)th fingertip center point and the approach vector} $\pv_i-\phi_p^{i}(\qv)$. Here $\pv_{i}$ serves as an auxiliary decision variable defining a point between the fingertip and object as in~\cite{mordatch-siggraph2012-cio}. \rebuttal{We initialize it halfway between the $i$th fingertip and estimated object surface.} We found that including $\pv_i$ was important for numerical stability. Finally, $\alpha$ weighs between minimizing distance and aligning orientation of the fingertip. We describe implementation details of $\sdfhat(\cdot)$ later in this section.

We solve the objective from Eq.~(\ref{eq:impl-obj-f}) using a Gauss-Newton solver with bound constrains to handle the joint limits with \rebuttal{$\alpha=\frac{1}{j^2}$. Here $j$ defines the current solver iteration}. We further implement a projection method on our line search to handle self collisions of the hand.
We show an example of a solution in Fig.~1 center left. We can see that the planned configuration makes contact with the object at the estimated object surface. However, the estimated object surface is likely inaccurate due to error in object pose estimation and shape reconstruction. In the next subsection, we describe how we execute our planned grasp to handle this uncertainty.

\subsection{Making Contact}
To move to the target configuration, we compute time-optimal
trajectories for each finger \cite{kunz-rss12-time-optimal} and then
scale the trajectories to be of equal duration, so
that all fingers reach their target configurations at the same time. This lowers the chance of moving the object when making
contact. The robot evaluates the constraint described in Eq.~(\ref{eq:sdf-constraint})
online using tactile feedback during trajectory execution. This allows the robot to detect contact before the
trajectory is completely executed. If the tactile classifier
$t(\tau[k])$ detects contact at timestep \(k\), the constraint is
satisfied and the finger in contact stops moving. \rebuttal{If after $K$ steps a finger reaches the desired configuration, but does not detect contact, the robot moves its finger toward the object surface following the gradient of $\sdfhat$ in attempt to make contact. The finger stops following the \(\sdfhat{}\) gradient if it hasn't made contact after \(\bar{k}\) additional control iterations as this implies the shape estimate gradient is uninformative at this location. Formally,	\(\pi(\qv[k], \tau[k])=\)
\begin{align}
	\begin{cases} 0 &\mbox{if}\, t(\tau[k]) \geq  p_c \vee k > K + \bar{k}\\
                \Delta \qv[k] &\mbox{if}\, t(\tau[k]) <  p_c \land k \leq K\\
	        -\gamma \nabla_{\qv} \sdfhat(\phi_p(\qv),\ov) &\mbox{if}\, t(\tau[k]) <  p_c \land K < k \leq K+\bar{k} 
              \end{cases}\raisetag{28pt}
\label{eq:tactile_controller}
\end{align} 
where \(\Delta \qv[k]\) is the initial plan of $K$ steps; \(p_{c}\) is the tactile pressure threshold of 365 \texttt{Pa}; \(\gamma=1.0\); and \(\bar{k} = 1\).}

The hand tracks \rebuttal{the joint positions commanded by $\pi$} using an inverse-dynamics torque controller. An example of a grasp after running our controller can be seen in Fig.~1 center right. Details about the classifier $t(\cdot)$ are explained in the next subsection. After making contact, we increase grasp force prior to lifting using a heuristic. If more force is not applied, the object slips through while the arm is attempting to lift it. The heuristic commands $0.7$ \texttt{rad} increase to all but the first joint on non-thumb fingers, and $0.5$ \texttt{rad} increase to last two joints of the thumb. A promising direction for future work is a more systematic reasoning about the applied force.

Despite making contact with the object surface, the resulting configuration might not generate a successful grasp. To estimate whether the robot should lift the object with the resulting configuration, we evaluate our in-contact grasp classifier $h_{c}$ to estimate the probability of grasp success. When evaluating $h_{c}$, we assume the object pose has not changed, despite applying more force, and feed in the same visualize input as for the preshape classifier. In-hand object pose tracking is a challenging problem and out of the scope for this paper. Similar to previous works~\cite{calandra-corl2017-feeling-of-success,wu-corl2019-mat}, if the output of the in-contact grasp classifier is below a threshold, the robot can reject the grasp and attempt a regrasp.

\subsection{Learned Model Architectures}
We now describe the learned MDN \(g(\cdot)\) and the preshape grasp classifier \(h_{p}(\cdot, \cdot)\), which are used to generate the preshape solution $\qv_0$. Then we describe the model used to estimate the initial object shape $\sdfhat(\cdot)$ and the tactile classifier \(t(\cdot)\) used to detect contact with the object. Finally, we describe our in-contact grasp classifier \(h_{c}(\cdot, \cdot)\) which is used to decide whether to lift the object or attempt a regrasp.

Following~\cite{lu-ram2020-MultiFingeredGP}, we use a mixture density network (MDN) $g(\cdot)$ that takes the voxelized object representation as an input and outputs a palm pose in object frame, as well as the joint positions for the two joints closest to the palm for each finger.
We model the preshape grasp classifier $h_{p}(\cdot, \cdot)$ as a neural network described in~\cite{lu-ram2020-MultiFingeredGP}. The classifier receives as input the palm pose in object frame, the voxelized object representation, the object dimensions and the first two joint values of every finger. The classifier outputs estimated probability of grasp success. Further details on the training procedure are in Section~\ref{sec:experiments}. We model our in-contact grasp classifier $h_{c}$ similarly to our preshape grasp classifier $h_{p}$ with one difference: $h_{c}$ receives all hand joints as input, while $h_{p}$ receives only two joints per finger.

To reconstruct object shape from a single view, we use the PointSDF model from~\cite{vandermerwe-icra2020-reconstruction-grasping} \rebuttal{as our implementation of $\sdfhat$}. The model takes as input the object point cloud and a query point and outputs the signed distance to the query point, \rebuttal{scaled to between -1 and 1.}
For further details see~\cite{vandermerwe-icra2020-reconstruction-grasping}.

We use the BioTac sensor installed on all fingertips for tactile feedback~\cite{wettels2014multimodal}. We define a contact classifier $t(\cdot)$ that receives as input the fluid pressure in the sensor ($P_{DC}$). Before the hand starts to close, the robot tares the sensor, such that the current pressure value reads zero. When closing the hand, if the pressure exceeds the preset threshold \(p_{c}\) for 10 consecutive timestamps, the classifier $t(\cdot)$ reports contact. The sensor is sampled at a frequency of 100 \texttt{Hz}.

\section{Experiments}
\label{sec:experiments}
We start this section by describing the data collection procedure used to train the grasp classifier.
\rebuttal{We then explain our baselines for comparison before discussing experiments and results for evaluating making contact and grasp success.}
Our experiments show: (1) that our proposed approach outperforms the state-of-the-art (SoTA) comparison as well as tactile enhanced SoTA; (2) the benefits of having both preshape and in-hand classifiers; and (3) the importance of estimating object geometry through the learned SDF.
Figure~\ref{fig:resulting-grasps} shows example grasps generated by our approach.

\subsection{Simulation: Data Collection and Training}
We conduct all training only in simulation using the four-fingered, $m=16$ DoF Allegro hand mounted on a Kuka LBR4 $n=7$ DoF arm, which is the same as our real world setup. We use a Microsoft Azure Kinect camera to generate the point cloud of the object on the table. We collected simulated grasp data using our robot hand-arm setup inside the Gazebo simulator with the DART physics engine. We use the built-in Gazebo Kinect camera to generate point clouds simulating the Microsoft Azure Kinect camera we use in real-world experiments.

We collected training data using a heuristic, geometry-based planner adapted from \cite{lu-ram2020-MultiFingeredGP}. However, our approach differs in how the robot closes its hand after moving to the preshape. We formulate hand-closing problem as the optimization problem defined in Eq.~(\ref{eq:impl-obj-f}), where we set $\alpha=0.0$ to speed up the optimization, and solve it using Gauss-Newton. After moving to the computed joint position, we increase the stiffness of the grasp by $0.2$\texttt{rad} to mimic a grasp stability controller. During data collection for training (and only for training), we use the true mesh of the object and no tactile feedback. To check for self-collisions, we use the GJK algorithm \cite{gjk}. In total, we collected 13,831 samples, out of which 3777 are successful grasps.

We pretrain a voxel autoencoder on a 3D reconstruction task~\cite{lu-ram2020-MultiFingeredGP} and extract the voxel encoder from the network for use in the grasp classifiers $h_{p}$ and $h_{c}$. We freeze the extracted weights and train the rest of the classifier using the Adam optimizer. We find that pretraining the autoencoder was a necessary step to train the classifiers successfully. On the validation set (10\% of the collected dataset) $h_{p}$ and $h_{c}$ achieve 78\% and 79\% accuracy respectively. This is comparable to numbers reported in \cite{lu-ram2020-MultiFingeredGP}. Following \cite{lu-ram2020-MultiFingeredGP}, we fit a mixture density network (MDN) that takes the object point cloud as input and outputs a grap preshape sample. The MDN uses the same voxel encoder as the classifiers.

\begin{figure}[h]
\centering
\includegraphics[width=0.49\textwidth]{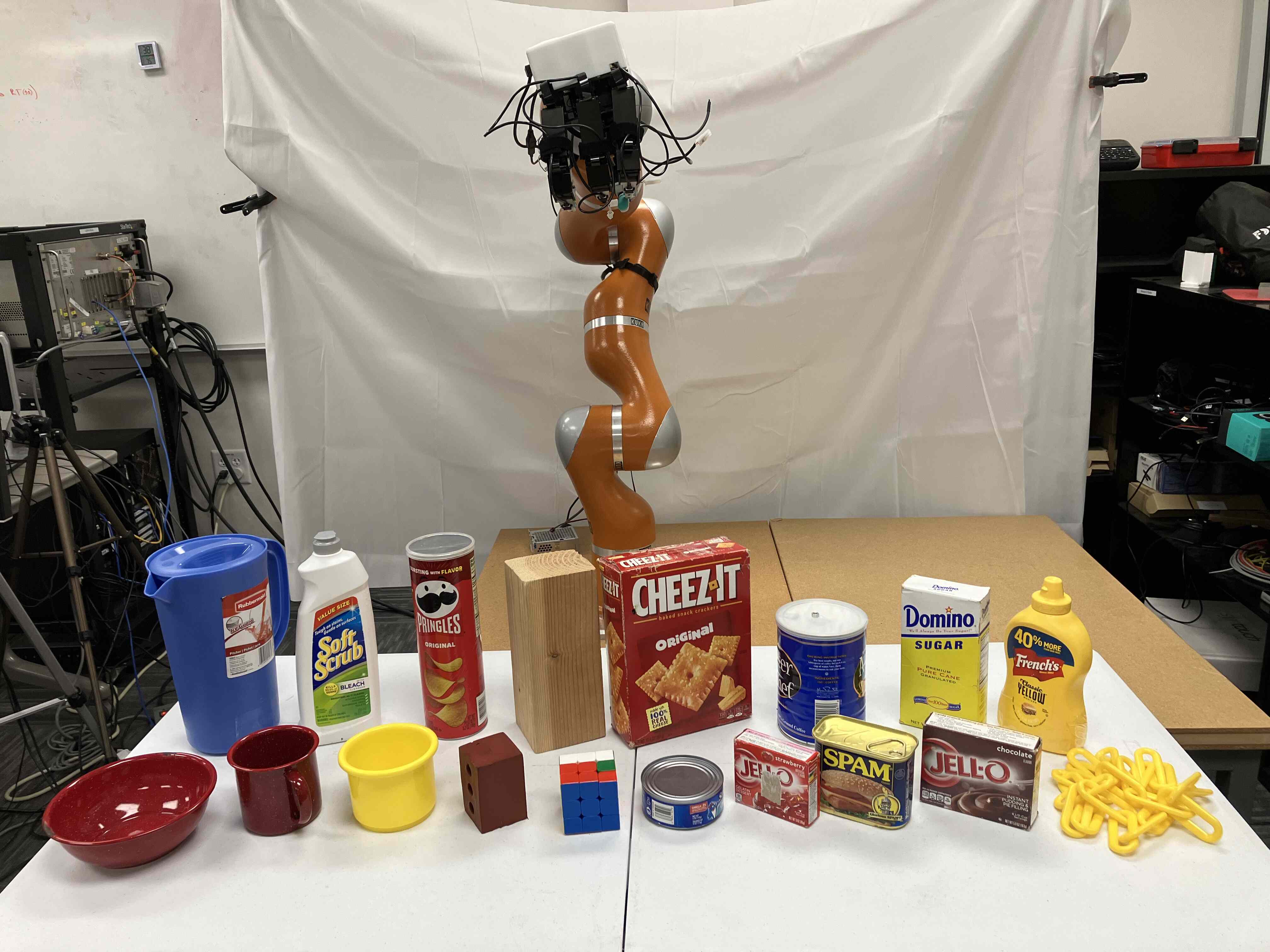}
\caption{The objects used in experiments. Objects from left to right; back row: ``pitcher", ``soft scrub", ``pringles", ``wooden block", ``cheezeit", ``master chef", ``domino", ``mustard"; front row: ``bowl", ``red cup", ``yellow cup", ``squishy brick", ``rubic cube", ``tuna can", ``jello red", ``spam", ``jello brown" and ``chain".}
\label{fig:experimental-setup}\vspace{-20pt}
\end{figure}

\subsection{Baselines for Comparison}
We implement \rebuttal{five} baseline comparisons to quantify the benefits of the different components of our approach. Our first baseline, \textbf{B1} uses the heuristic, grasp controller from~\cite{lu-ral2019-grasp-type} to close the hand. This represents a SoTA comparison to our complete system.
Since \textbf{B1} does not use tactile sensors, we implement \textbf{B2} that uses the same controller as \textbf{B1}, but stops when the tactile classifier detects contact instead of the sensed joint velocities.
This allows us to understand to what extent any improvement comes from adding tactile sensors to the grasp controller.
Baseline \textbf{B3} samples a preshape configuration from an MDN trained on all positive samples from the training dataset, but closes the hand using our visual-tactile controller. This examines whether our controller removes the need for the initial preshape optimization and hence the preshape classifier.
Further, we look into the importance of object shape completion by introducing \textbf{B4} which reconstructs only the visible region of the object using marching cubes~\cite{mcubes} from the camera's partial point cloud, which we use with the trimesh library~\cite{trimesh} to evaluate the SDF. This contrasts with our approach which uses PointSDF to estimate the entire shape of the object.
\rebuttal{We implement baseline \textbf{B5} as an additional comparison for evaluating the ability for the controllers to make fingertip contact with the object without moving it. We simply execute our controller \(\pi\) open-loop stopping after \(K\) steps ignoring tactile feedback.}

\subsection{Making Contact in the Real World}
\rebuttal{We first evaluate our proposed controller on generating grasps that make fingertip contact with the object without moving it.
  We report the number of fingertips in contact with the object after closing the hand, as well as whether the object remained flat on the table, tilted, or was knocked over by the robot. We compare our controller to \textbf{B1} and \textbf{B5} by generating and saving a preshape for a given object using our grasp planner and then executing the selected grasp controller. We manually label whether the object tilted or fell. We run experiments on 5 different objects (pringles, domino, soft scrub, mustard, and wooden block) in 5 different poses resulting in 75 hand closing episodes in total.}\vspace{-12pt}
\begin{figure}[h]
\centering
\includegraphics[width=0.5\textwidth]{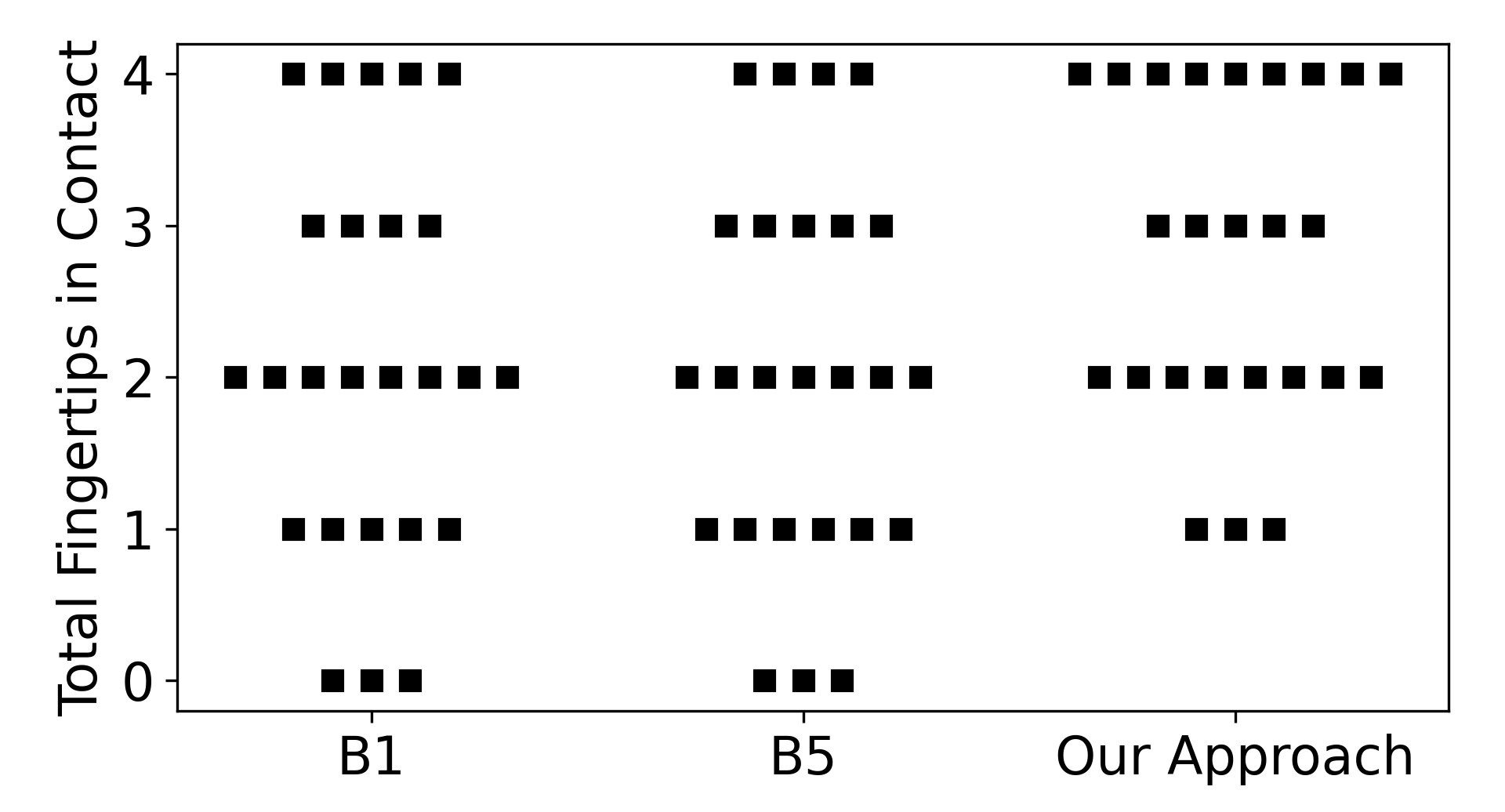}
	\caption{Number of fingertips in contact with the object after closing.}
\label{fig:making-contact}\vspace{-12pt}
\end{figure}

\rebuttal{Figure~\ref{fig:making-contact} visualizes our results. Our approach results in more fingertips in contact with the object than using \textbf{B1} or \textbf{B5}. The object tilted only 3 times when using our controller, while \textbf{B5} tilted the object 10 times, and \textbf{B1} tilted 21 times. Further, we report that our method never knocked over the object, while the vision-only \textbf{B5} knocked over an object once, and \textbf{B1} did so three times.
  Our method sometimes failed in making all four fingers contact the object. We believe three issues contribute to this: (1) the object simply stands out of reach for some fingers for the generated preshape, (2) the $\sdfhat$ is not accurate enough to lead the controller in the right direction, (3) the tactile contact classifier is sometimes inaccurate. While imperfect, we show next that our approach provides advantages over the comparison methods for our task of interest--grasping.}

\subsection{Grasping in the Real World}
We perform real-world experiments using 18 objects from the YCB~\cite{ycb} object set. These objects span different textures, shapes, masses, and sizes. We show the experimental setup and objects used in Fig.~\ref{fig:experimental-setup}. All experimental objects but the pringles can are unseen in training. For experiments, we place a single target object on the table and run our pipeline. We chose a fixed test location on the table for the target object and rotate the object to 3 different angles, keeping the poses constant across the methods and objects. Given a resulting robot joint configuration from our grasp planner, we perform motion planning using the RRTConnect planner from MoveIt! to obtain a collision free trajectory to the preshape configuration. We provide the bounding box of the object to the motion planner for collision checking. We find that this conservative approach yields fewer collisions with the object compared to providing the reconstructed mesh of the object to the motion planner. After executing the motion plan, the robot closes its hand using the grasp controller. Then, the robot attempts to lift the object approximately 15cm above the table. Only if the object is in the hand after executing the lifting trajectory, we label the attempt as a successful grasp. If the robot moved to the preshape, but failed to lift the object, we label the attempt as unsuccessful. \rebuttal{The robot attempts 270 grasps total, 54 for each of the 5 comparative methods.}\vspace{-2pt}

\begin{figure}[h]
\centering
	\includegraphics[width=0.2\textwidth, angle=270]{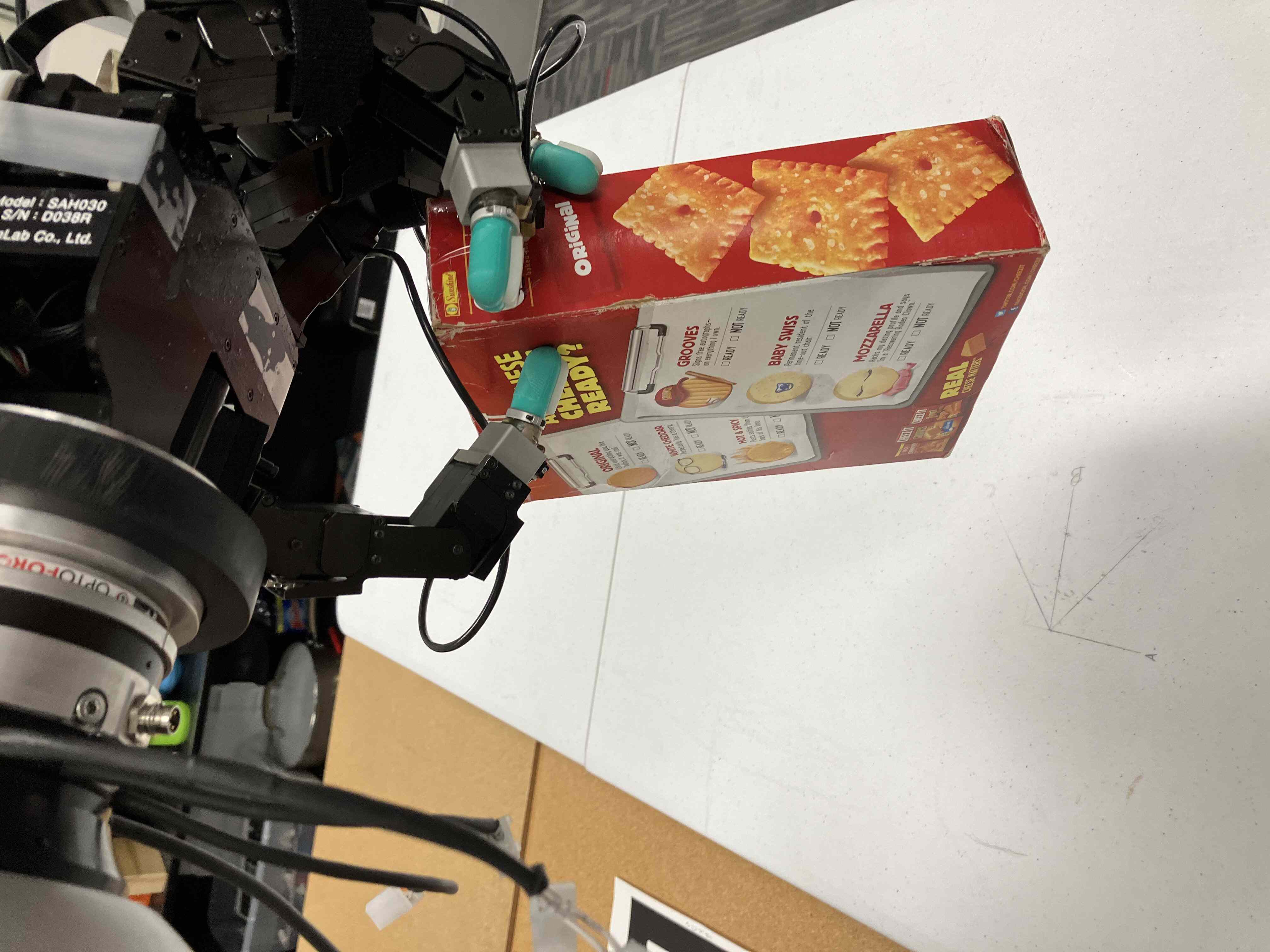}
	\includegraphics[width=0.2\textwidth, angle=270]{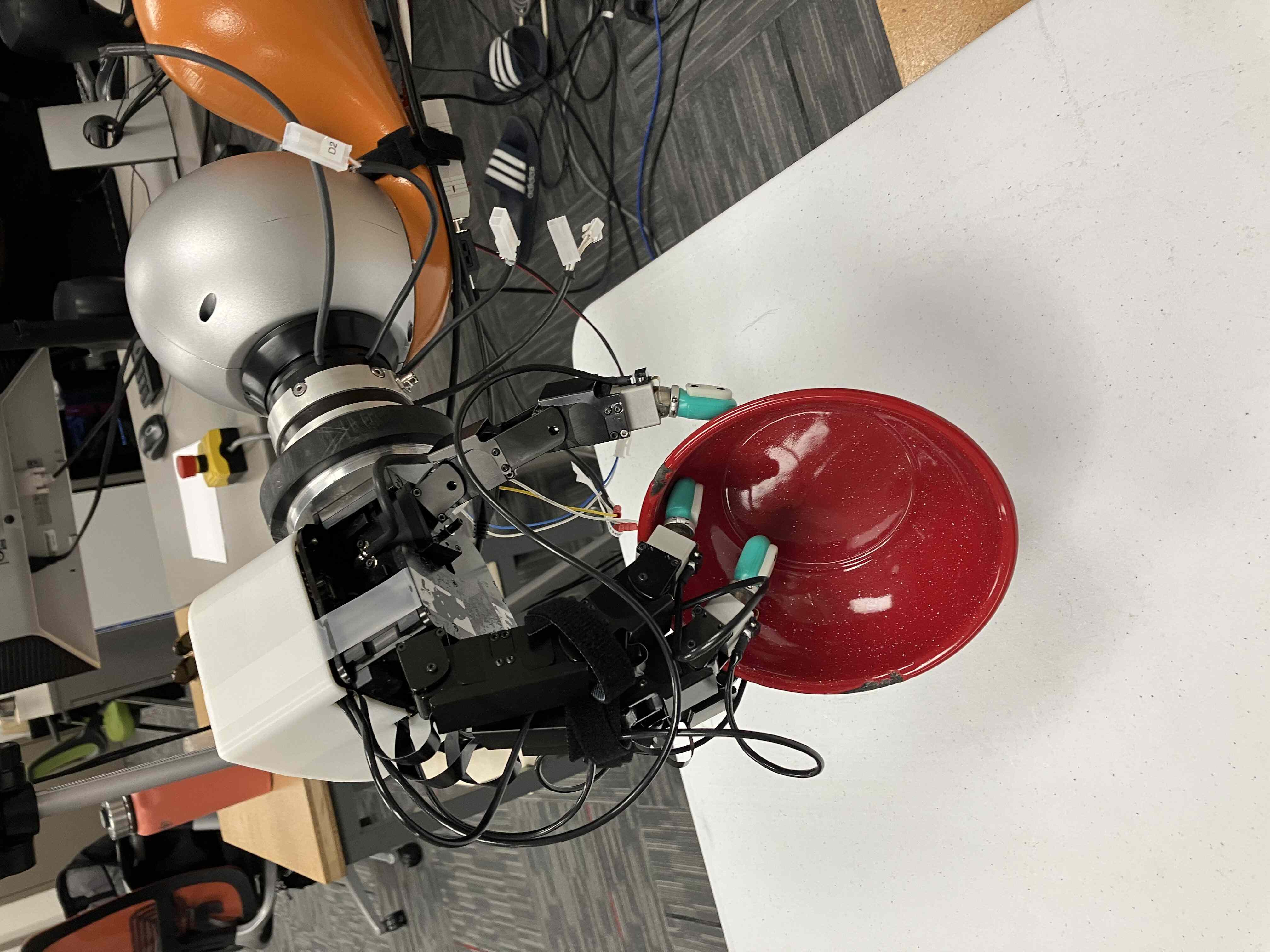}
	\includegraphics[width=0.2\textwidth, angle=270]{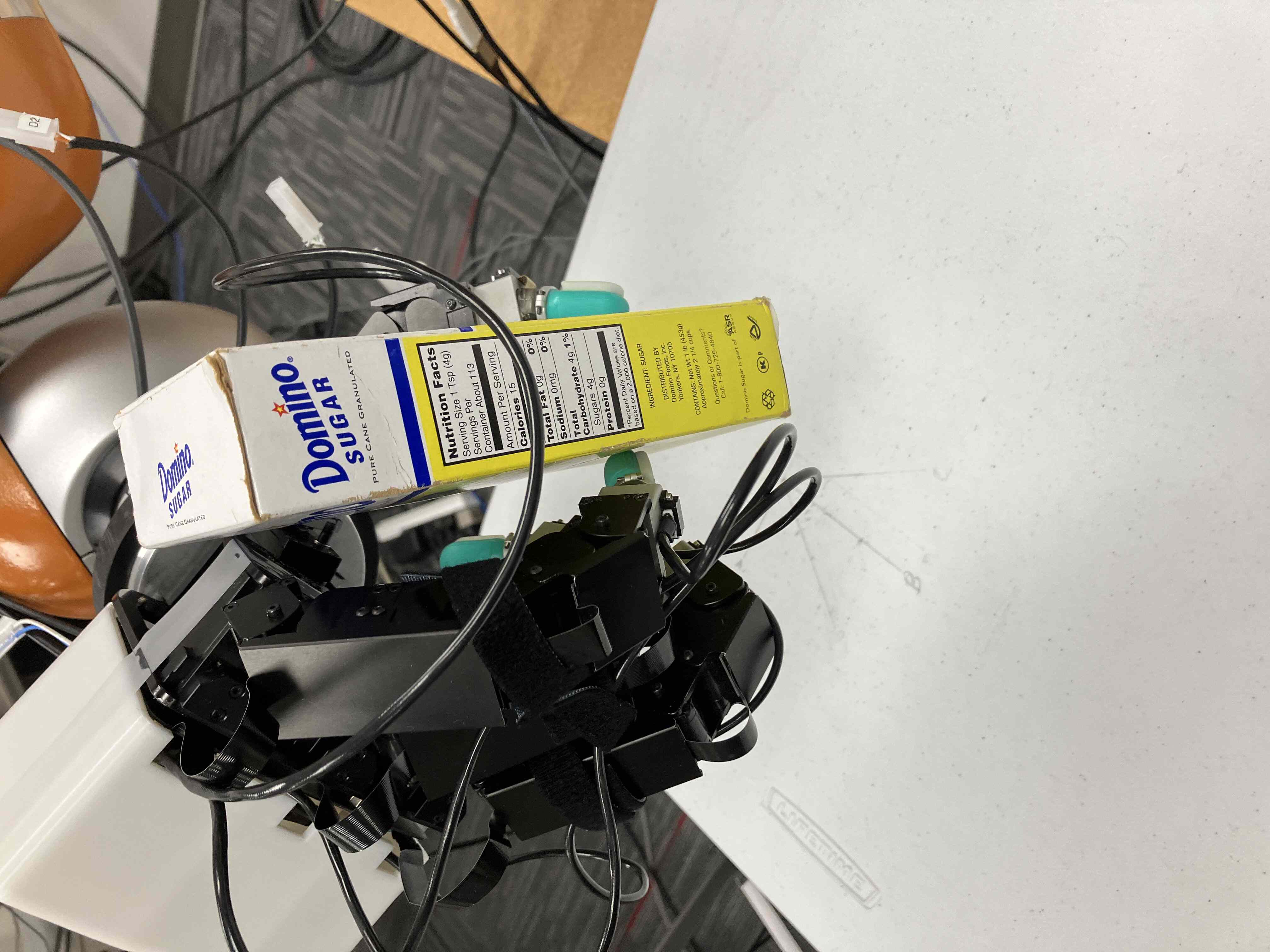}

	\includegraphics[width=0.2\textwidth, angle=270]{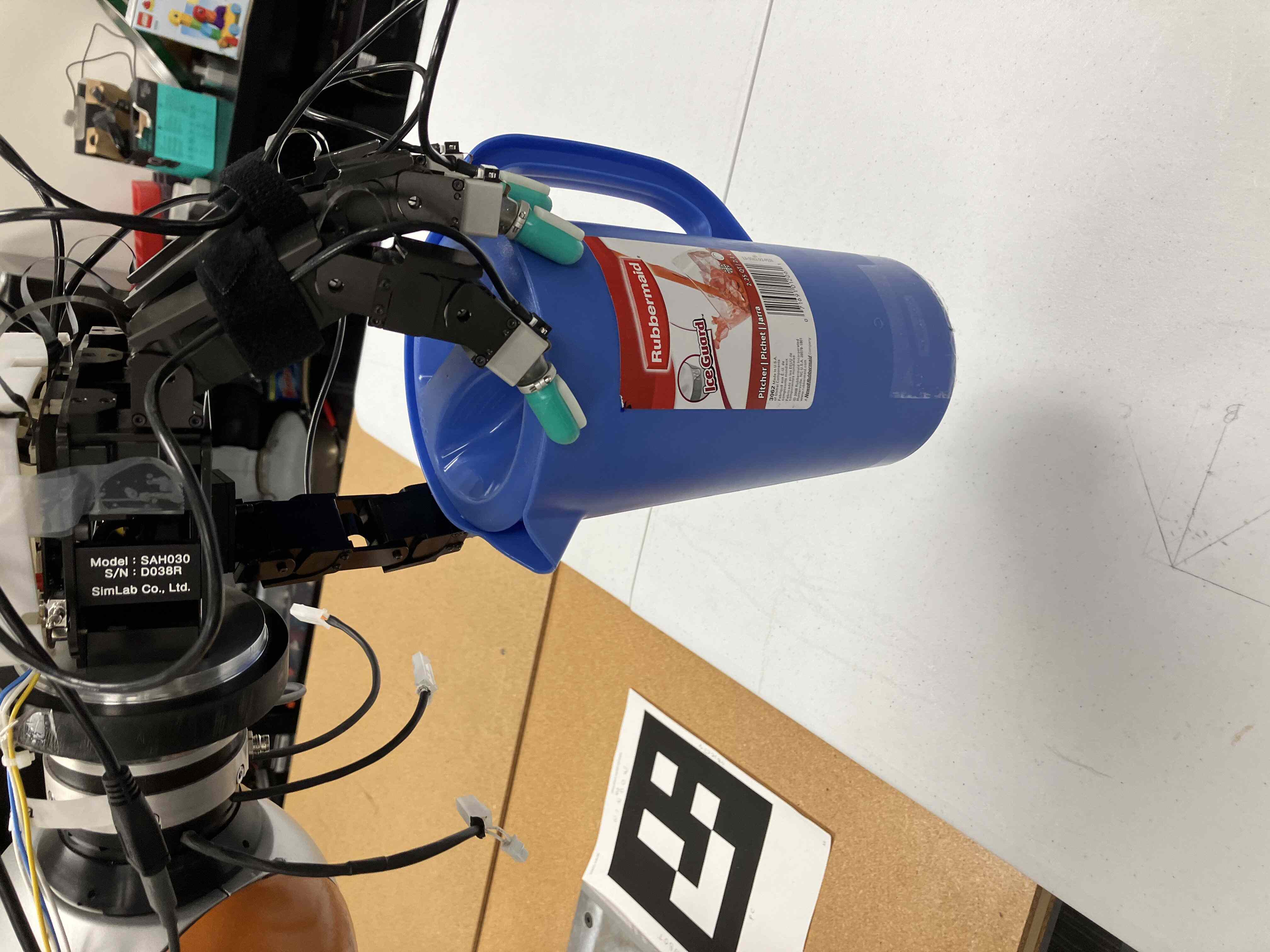}
	\includegraphics[width=0.2\textwidth, angle=270]{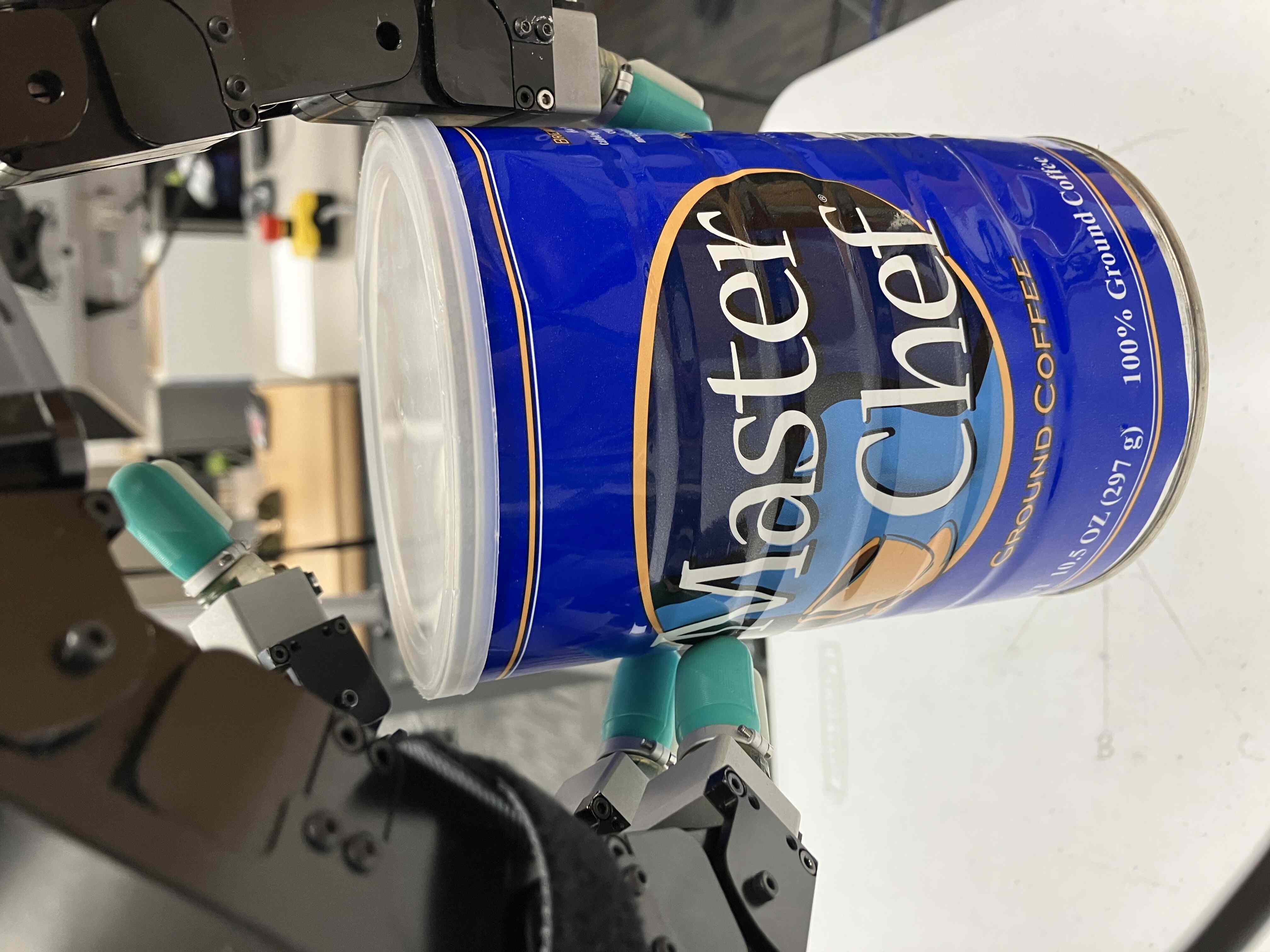}
	\includegraphics[width=0.2\textwidth, angle=270]{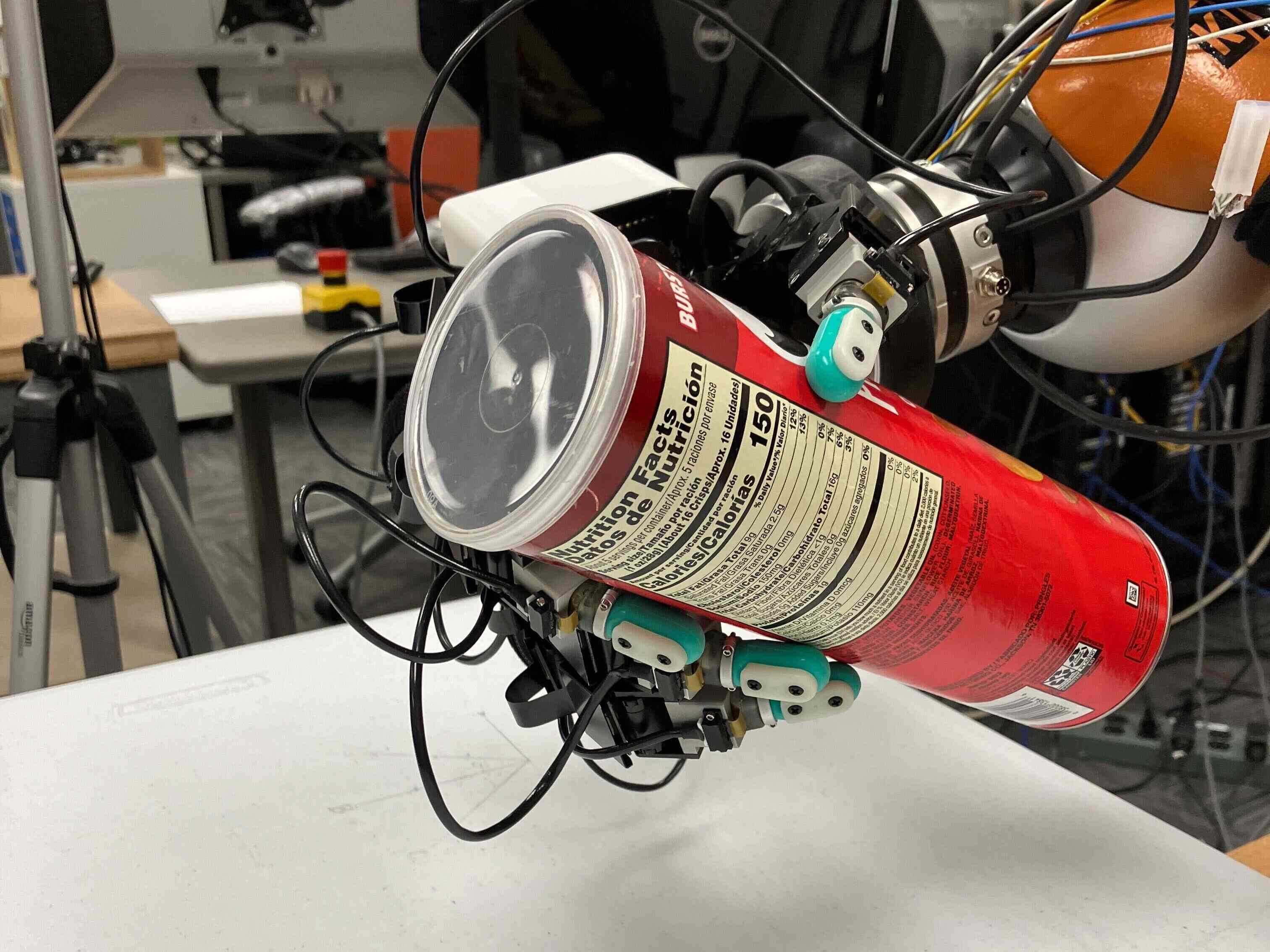}

	\includegraphics[width=0.2\textwidth, angle=270]{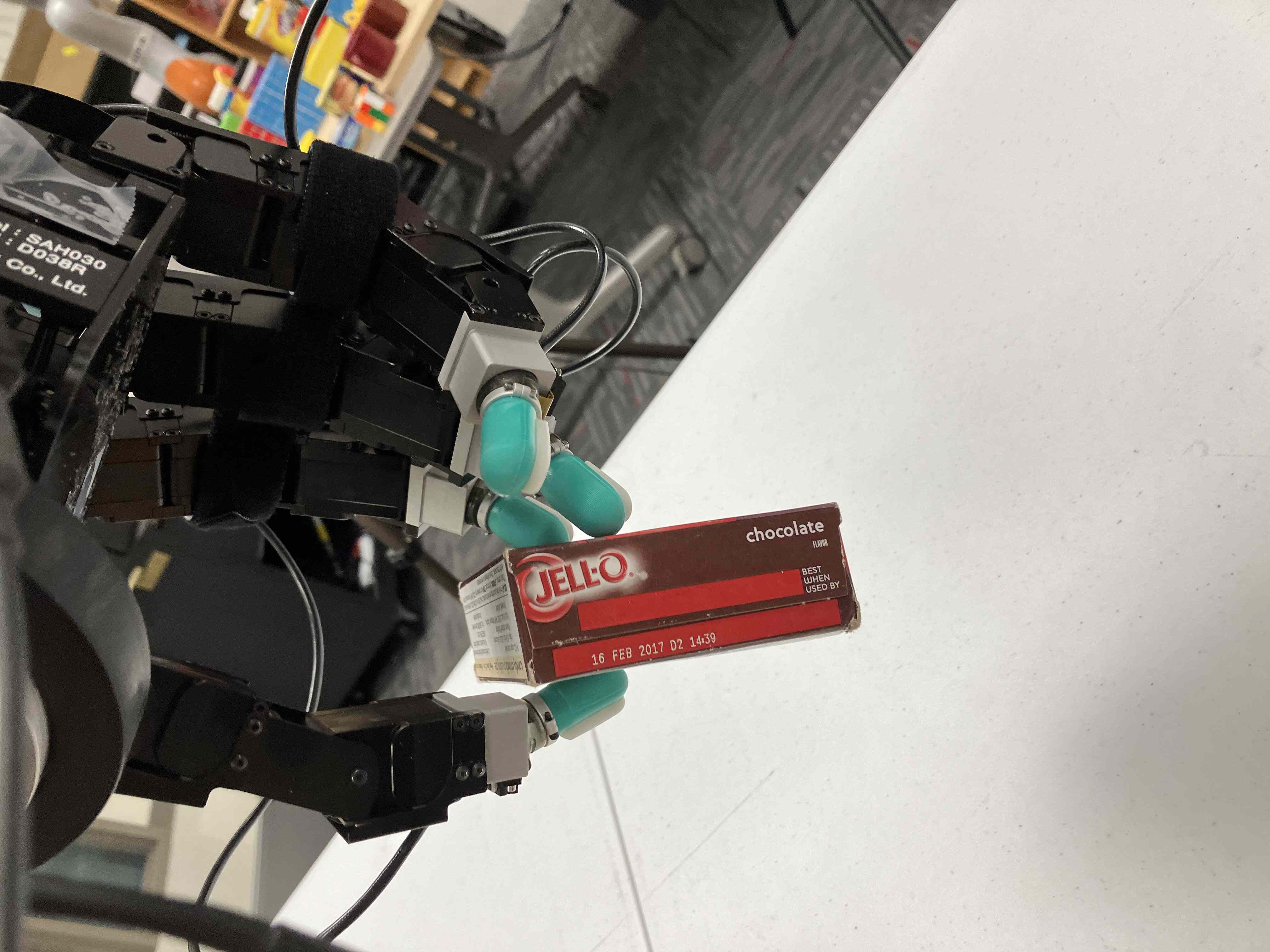}
	\includegraphics[width=0.2\textwidth, angle=270]{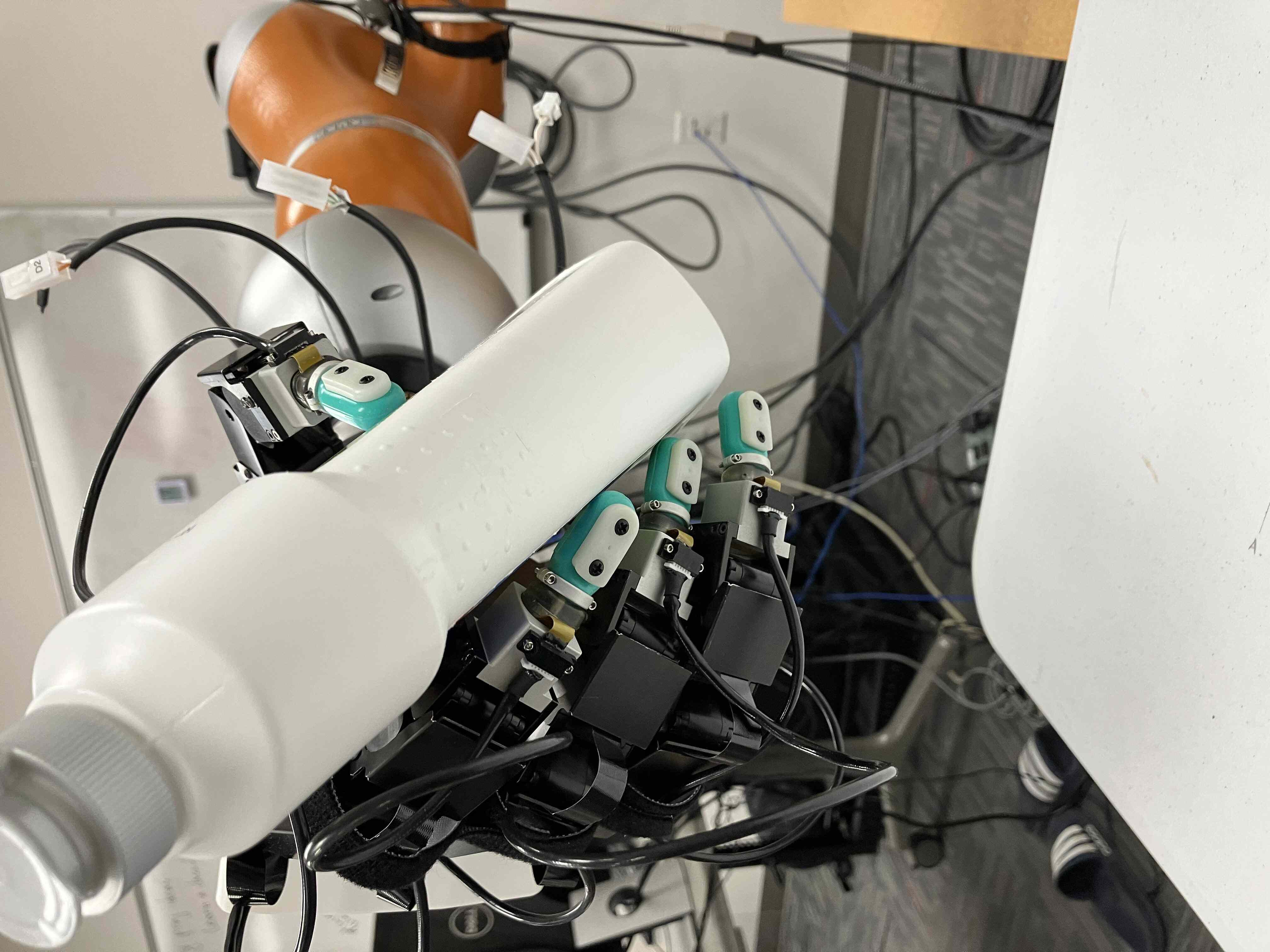}
	\includegraphics[width=0.2\textwidth, angle=270]{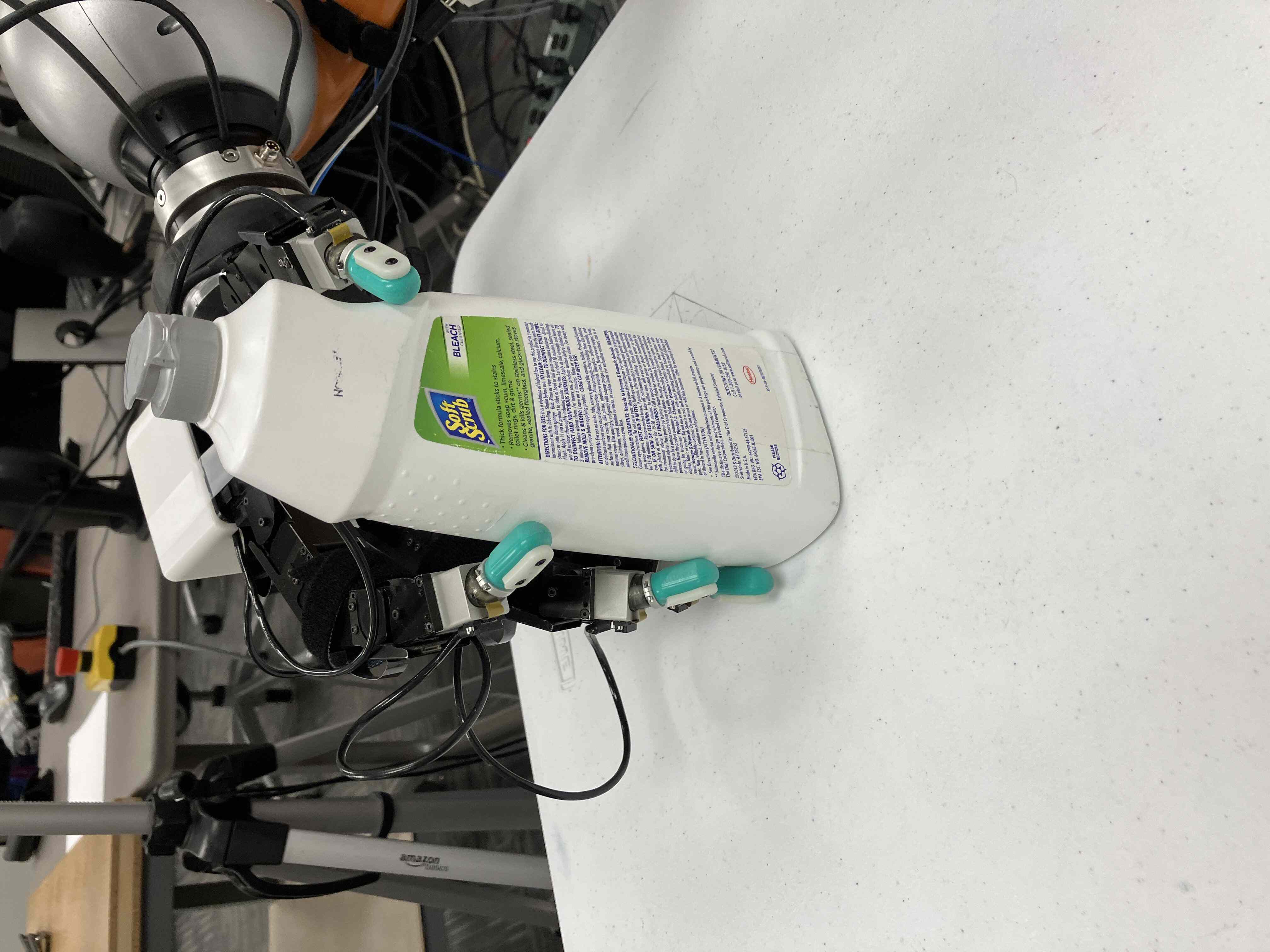}
	\includegraphics[width=0.2\textwidth, angle=270]{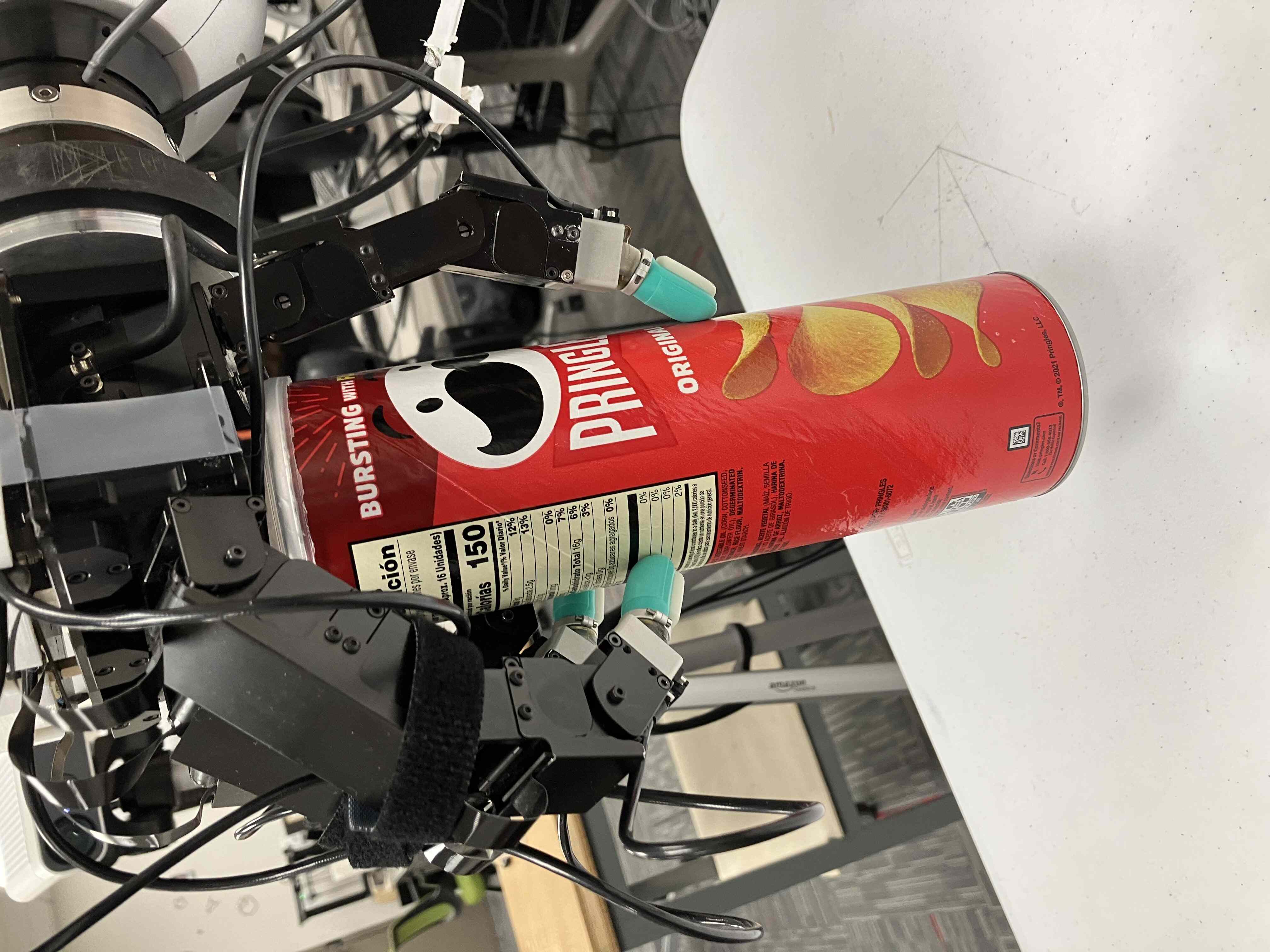}
	\includegraphics[width=0.2\textwidth, angle=270]{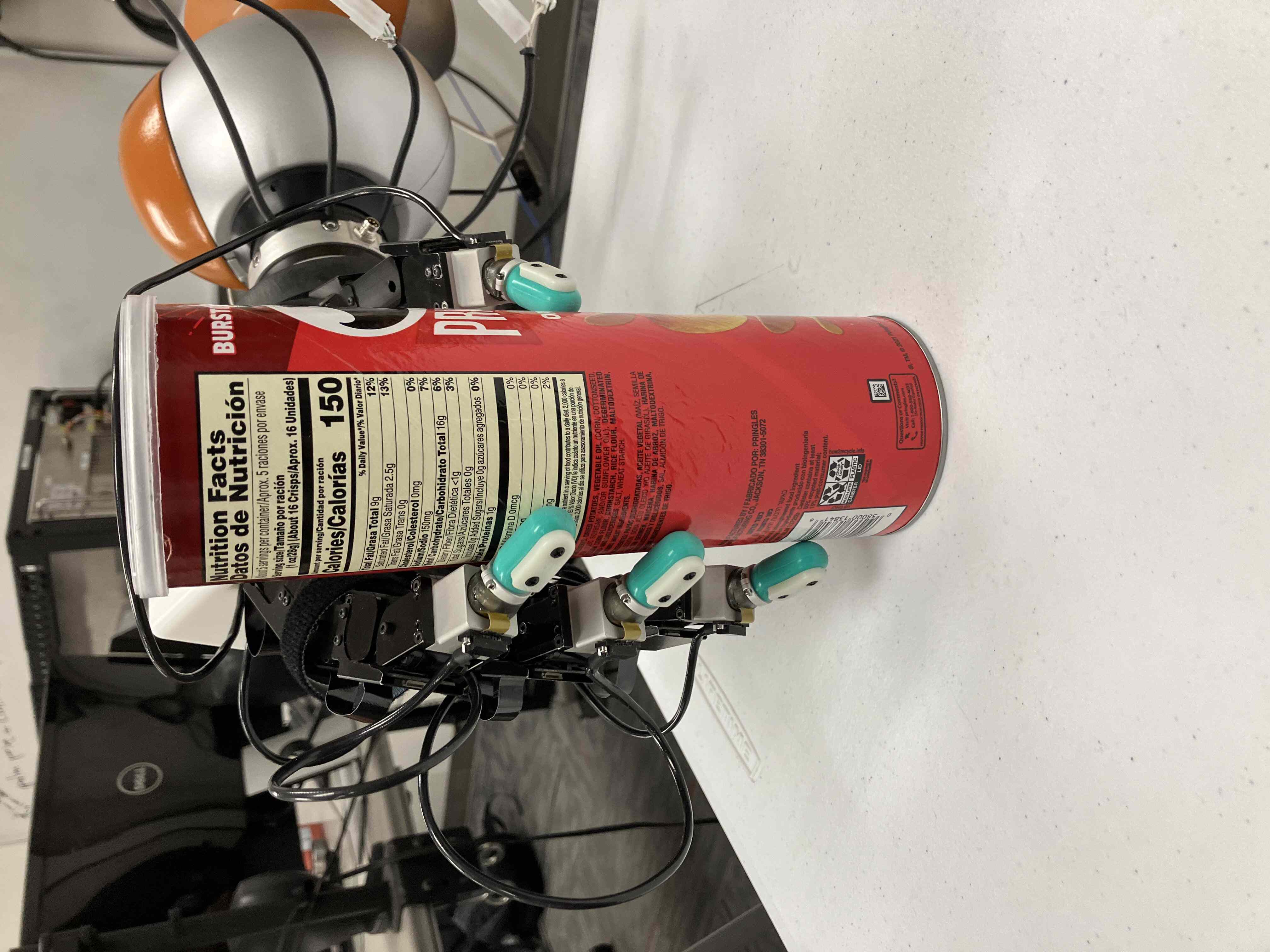}
	\includegraphics[width=0.2\textwidth, angle=270]{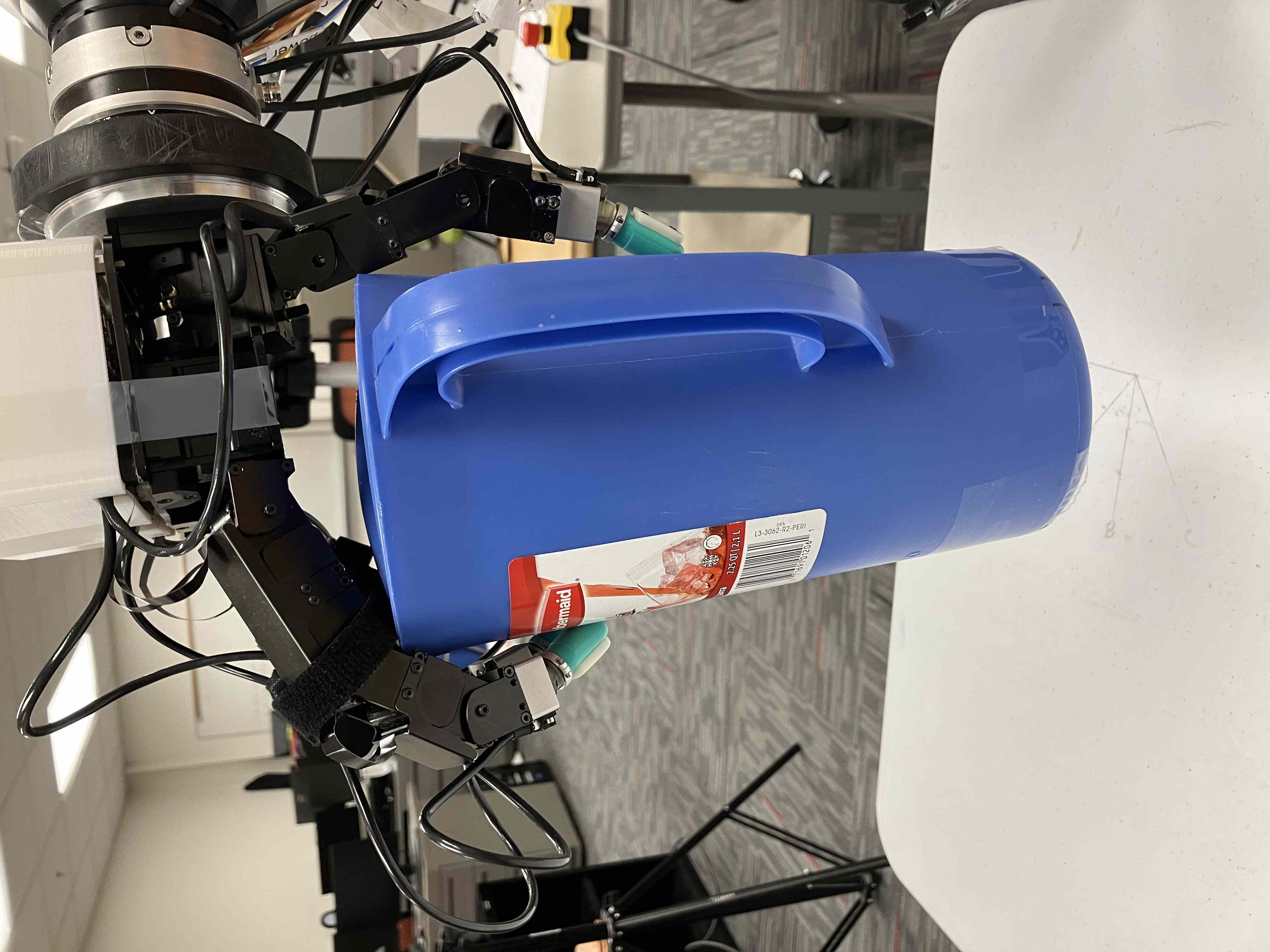}
	\includegraphics[width=0.2\textwidth, angle=270]{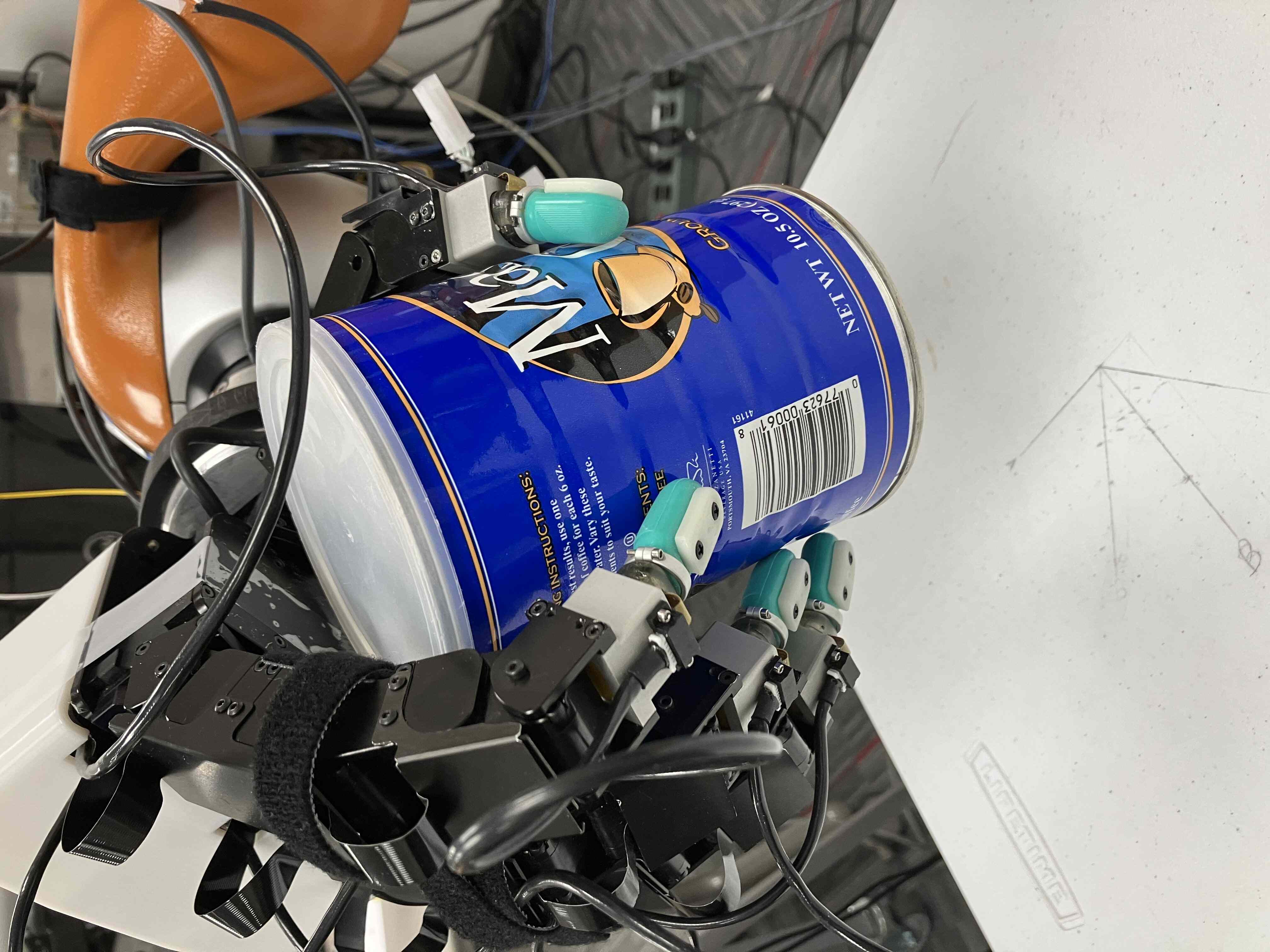}
	\includegraphics[width=0.2\textwidth, angle=270]{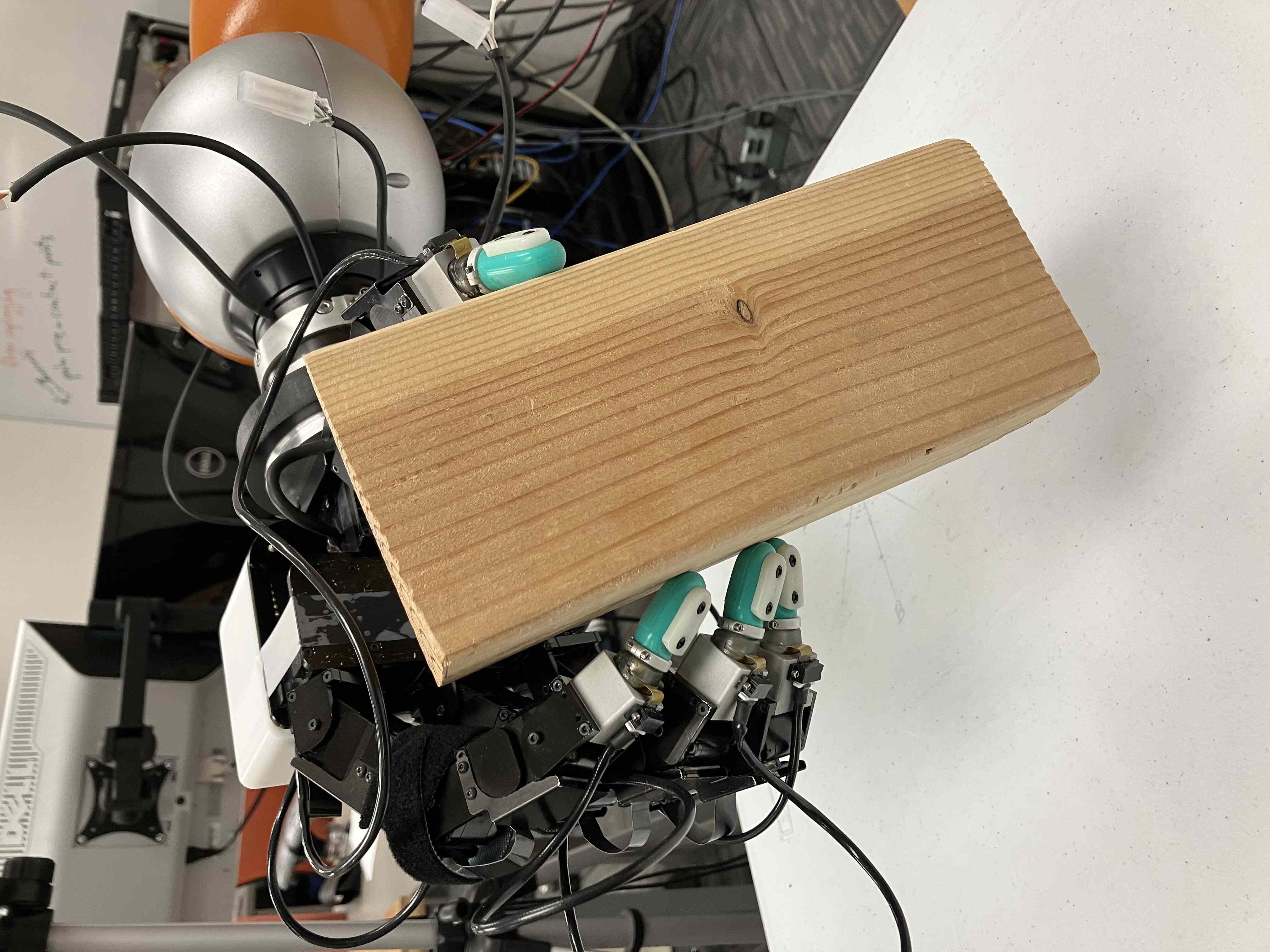}
	\includegraphics[width=0.2\textwidth, angle=270]{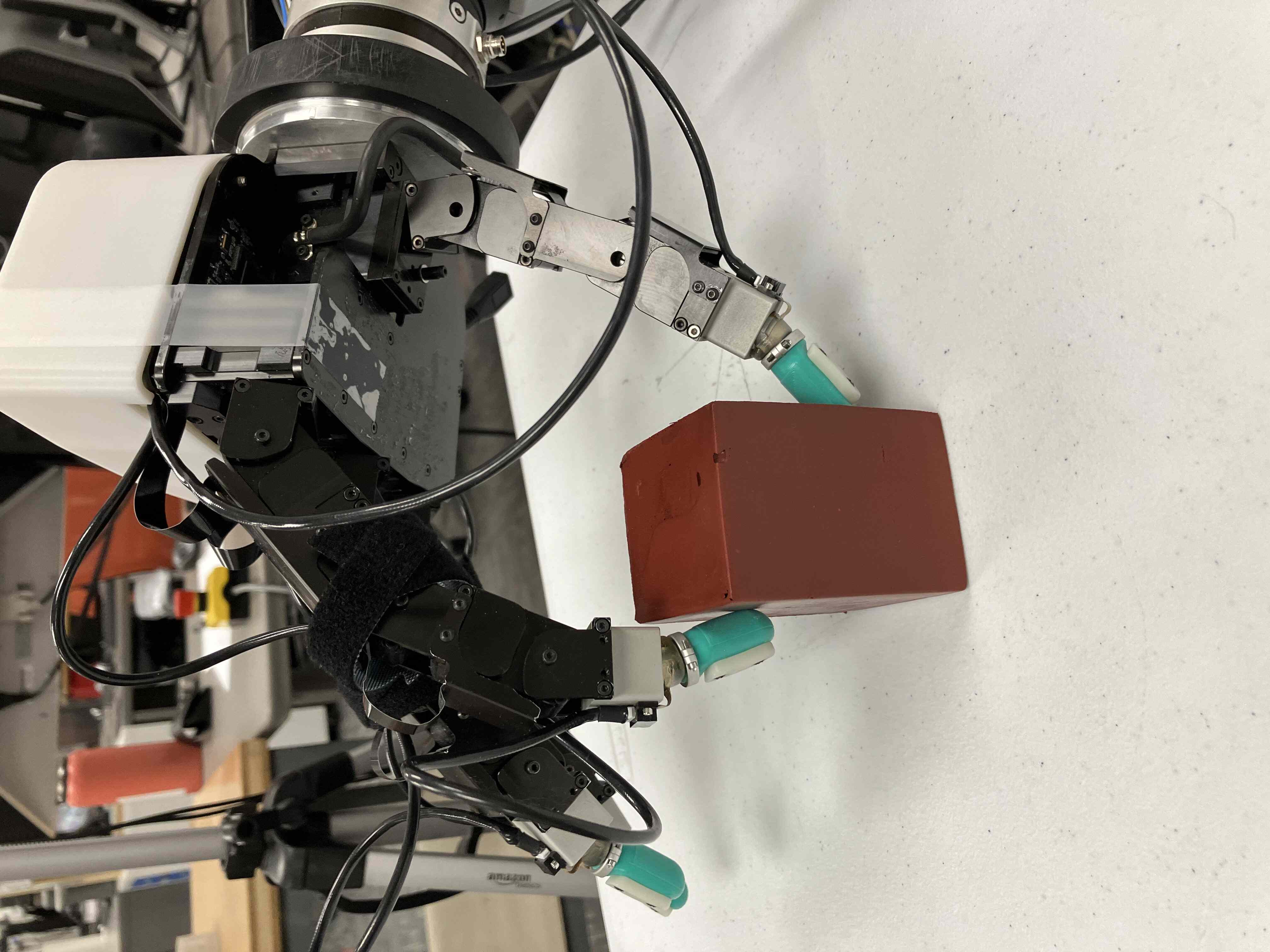}
	\caption{Example precision grasps resulting from our controller.}
\label{fig:resulting-grasps}\vspace{-20pt}
\end{figure}

\subsubsection{Quantitative results}
\begin{figure}[H]\vspace{-5pt}
\centering
\includegraphics[width=\linewidth, trim={0.4cm 0.5cm 0.4cm 2.2cm}, clip]{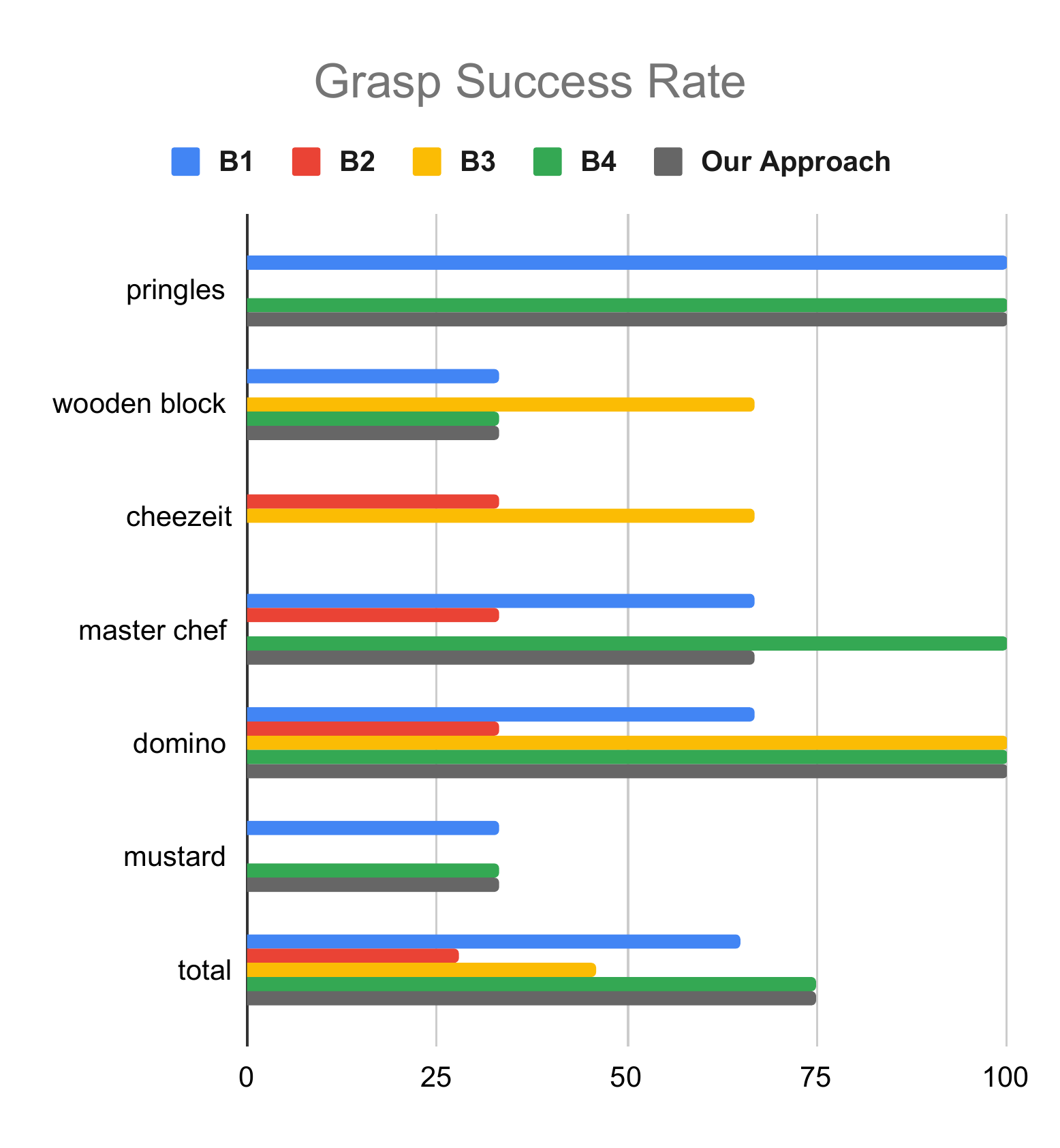}
\caption{Grasp success rates for each tall object and overall.}
\label{fig:plot_success}\vspace{-12pt}
\end{figure}

Figure~\ref{fig:plot_success} presents the grasp success rates for the taller objects that comprise the back row in Fig.~\ref{fig:experimental-setup}.
Overall success rates for \textbf{B1-4} and our approach are $65\%, 28\%, 46\%, 75\%$, and $75\%$, respectively. 

Our approach outperforms most of the baselines and successfully picks up most of the objects. However, \textbf{B1}, \textbf{B4}, and our approach all fail to find a plan to pick up ``cheezeit''.
This arises from the initial grasp preshape optimization always coming into contact with the object. This seems to be somewhat an issue of random initialization as \textbf{B2}, which uses the same preshape generation was successful once. However, both \textbf{B2} and \textbf{B3} had the overall lowest performance.
Only \textbf{B4} has a success rate equal to our proposed approach.
As such we see that the proposed in-contact grasp planner and associated tactile feedback controller combine to improve over the previous SoTA \textbf{B1}. We can attribute the lack of improvement from using tactile feedback in \textbf{B2} to the fact that the heuristic grasp controller does not explicitly try to make contact with the tactile fingertips. Thus \textbf{B2} would often make contact with other parts of the hand when closing pushing away or knocking over the object. In contrast, while our in-contact planner would generate target configurations that penetrate the object, the resulting grasps often succeeded as the robot detected contact with the tactile sensors.

Recall that the only difference between our proposed approach and \textbf{B4} is the SDF estimate. To further investigate the difference in performance, we examine the ``low profile'' objects shown in the front row of Fig.~\ref{fig:experimental-setup}.
We define ``low profile'' objects to be objects shorter than $10$\texttt{cm}. The tallest low-profile object is ``jello brown'' with a height of $89$\texttt{mm}. 
Our approach achieves a \textbf{40\%} rate for low-profile objects, while \textbf{B1-4} have respective success rates of $20\%, 9\%, 10\%$ and $18\%$. This result demonstrates a substantial improvement in using PointSDF for reasoning about small-object geometry
compared with running marching cubes in $\textbf{B4}$ for planning.

That said, across the 10 different low profile objects, we were able to execute only 78 out of 150 attempts, because of the resulting preshape coming into collision with the table. This arises from the lack of explicit reasoning about collisions with the environment in our optimization. We point out that this is one of the biggest shortcomings of our method and we believe this is a promising avenue for future work.

\begin{figure}[h]\vspace{-10pt}
\centering
\includegraphics[width=0.45\textwidth, trim={0.4cm 0.4cm 0.2cm 0.2cm}, clip]{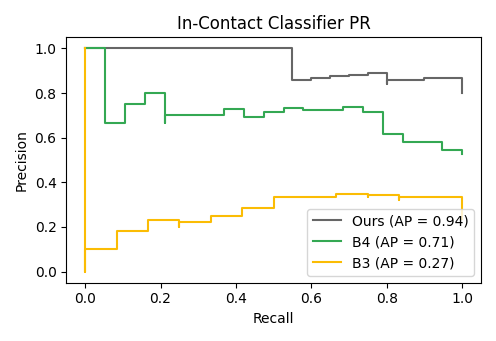}
\caption{Precision-recall curve for our in-contact grasp classifier, \(h_c\). We compare the performance of our approach, \textbf{B3}, and \textbf{B4}, the three approaches that use our proposed tactile controller.}
\label{fig:inhand_pr}\vspace{-10pt}
\end{figure}

Finally, we look into the performance of our in-contact grasp classifier $h_{c}$.
We analyze the performance of the classifier using only \textbf{B3, B4} and our method as those use our proposed grasp controller.
We filter out the data where the trajectory couldn't be executed.
We show the resulting precision-recall curve for executed grasps in Fig.~\ref{fig:inhand_pr}.
This shows the classifier performs well in the real world on previously unseen objects despite being trained only in simulation. It also shows that the classifier can be used to further increase grasp success rates by executing grasps only if the classifier prediction is above a high threshold.
To construct the precision-recall curves, in our experiments we didn't filter out any grasps based on the estimated probability of success.
Finally, we see that the $h_{c}$ obtains both high precision and recall when coupled with our proposed approach.

\subsubsection{Qualitative results}
While our approach outperforms the baselines quantitatively, we think it shines brightest when one looks at the generated grasps. In Fig.~\ref{fig:resulting-grasps} we show some grasps resulting from our controller, where the robot successfully grasps objects such that the tactile sensors contact the object. Since \textbf{B1} and \textbf{B2} don't explicitly reason about making contact with the fingertips, we found that the resulting grasps tend to make contact with other parts of the hand instead. This makes our approach suitable for downstream tasks that require updating the object shape or pose estimate or requiring precision in-hand manipulation.

\section{Conclusion}
\label{sec:conclusion}
We proposed an approach for a robot to plan and execute multi-fingered, precision grasps for previously unseen objects. Key to our methods success is the use of initial shape estimation from a single camera view as a constraint in the grasp planner, coupled with the use of tactile sensing to adapt the grasp online. Despite training only in simulation, our approach works well in the real world and generates more precision grasps than the comparison approach~\cite{lu-ram2020-MultiFingeredGP}.

\vspace{-5pt}

\bibliographystyle{IEEEtran}
{\footnotesize
\bibliography{references}

\begin{thebibliography}{10}
\providecommand{\url}[1]{#1}
\csname url@samestyle\endcsname
\providecommand{\newblock}{\relax}
\providecommand{\bibinfo}[2]{#2}
\providecommand{\BIBentrySTDinterwordspacing}{\spaceskip=0pt\relax}
\providecommand{\BIBentryALTinterwordstretchfactor}{4}
\providecommand{\BIBentryALTinterwordspacing}{\spaceskip=\fontdimen2\font plus
\BIBentryALTinterwordstretchfactor\fontdimen3\font minus
  \fontdimen4\font\relax}
\providecommand{\BIBforeignlanguage}[2]{{%
\expandafter\ifx\csname l@#1\endcsname\relax
\typeout{** WARNING: IEEEtran.bst: No hyphenation pattern has been}%
\typeout{** loaded for the language `#1'. Using the pattern for}%
\typeout{** the default language instead.}%
\else
\language=\csname l@#1\endcsname
\fi
#2}}
\providecommand{\BIBdecl}{\relax}
\BIBdecl

\bibitem{sundaralingam-ral2020-benchmarking-in-hand}
S.~Cruciani, B.~Sundaralingam, K.~Hang, V.~Kumar, T.~Hermans, and D.~Kragic,
  ``{Benchmarking In-Hand Manipulation},'' \emph{{IEEE Robotics and Automation
  Letters}}, vol.~5, no.~2, 1 2020.

\bibitem{lu-ral2019-grasp-type}
Q.~Lu and T.~Hermans, ``{Modeling Grasp Type Improves Learning-Based Grasp
  Planning},'' \emph{IEEE Robotics and Automation Letters}, 2019.

\bibitem{lu-iros2020-active-grasp}
Q.~Lu, M.~V. der Merwe, and T.~Hermans, ``{Multi-Fingered Active Grasp
  Learning},'' in \emph{IEEE/RSJ Intl. Conf. on Intelligent Robots and
  Systems}, 2020.

\bibitem{yamaguchi-humanoids2016-fingervision}
A.~Yamaguchi and C.~G. Atkeson, ``{Combining Finger Vision and Optical Tactile
  Sensing: Reducing and Handling Errors While Cutting Vegetables},'' in
  \emph{IEEE/RAS Intl. Conf. on Humanoid Robots}, 2016.

\bibitem{hogan-iros2018-tactile-regrasp}
F.~R. {Hogan}, M.~{Bauza}, O.~{Canal}, E.~{Donlon}, and A.~{Rodriguez},
  ``Tactile regrasp: Grasp adjustments via simulated tactile transformations,''
  in \emph{IEEE/RSJ Intl. Conf. on Intelligent Robots and Systems}, 2018.

\bibitem{bala-icra2019-tactile-force-estimation}
B.~Sundaralingam, A.~S. Lambert, A.~Handa, B.~Boots, T.~Hermans, S.~Birchfield,
  N.~Ratliff, and D.~Fox, ``Robust learning of tactile force estimation through
  robot interaction,'' in \emph{IEEE Intl. Conf. on Robotics and Automation},
  2019.

\bibitem{calandra-corl2017-feeling-of-success}
R.~Calandra, A.~Owens, M.~Upadhyaya, W.~Yuan, J.~Lin, E.~H. Adelson, and
  S.~Levine, ``The feeling of success: Does touch sensing help predict grasp
  outcomes?'' in \emph{Conference on Robot Learning}, 2017.

\bibitem{murali-iser2018-without-seeing}
A.~Murali, Y.~Li, D.~Gandhi, and A.~Gupta, ``Learning to grasp without
  seeing,'' in \emph{Intl. Symposium on Experimental Robotics}, 2018.

\bibitem{veiga-sensors2020-finger-control}
F.~Veiga, B.~B. Edin, and J.~Peters, ``In-hand object stabilization by
  independent finger control,'' \emph{Sensors}, vol.~20, no.~6, 2020.

\bibitem{bohg-tro2014-grasp-survey}
J.~Bohg, A.~Morales, T.~Asfour, and D.~Kragic, ``Data-driven grasp
  synthesis—a survey,'' \emph{{IEEE} Trans. on Robotics}, vol.~30, no.~2,
  2014.

\bibitem{vandermerwe-icra2020-reconstruction-grasping}
M.~Van~der Merwe, Q.~Lu, B.~Sundaralingam, M.~Matak, and T.~Hermans,
  ``{Learning Continuous 3D Reconstructions for Geometrically Aware
  Grasping},'' in \emph{IEEE Intl. Conf. on Robotics and Automation}, 2020.

\bibitem{lundell-2021icra-fingan}
J.~Lundell, E.~Corona, T.~Nguyen~Le, F.~Verdoja, P.~Weinzaepfel, G.~Rogez,
  F.~Moreno-Noguer, and V.~Kyrki, ``Multi-fingan: Generative coarse-to-fine
  sampling of multi-finger grasps,'' in \emph{IEEE Intl. Conf. on Robotics and
  Automation}, 2021.

\bibitem{jiang2021synergies}
Z.~Jiang, Y.~Zhu, M.~Svetlik, K.~Fang, and Y.~Zhu, ``Synergies between
  affordance and geometry: 6-dof grasp detection via implicit
  representations,'' in \emph{Robotics: Science and Systems}, 2021.

\bibitem{dragiev-icra2013-uncertainty-grasping-tactile}
S.~{Dragiev}, M.~{Toussaint}, and M.~{Gienger}, ``Uncertainty aware grasping
  and tactile exploration,'' in \emph{IEEE Intl. Conf. on Robotics and
  Automation}, 2013.

\bibitem{yi-iros2016-active-touch}
Z.~Yi, R.~Calandra, F.~Veiga, H.~van Hoof, T.~Hermans, Y.~Zhang, and J.~Peters,
  ``{Active Tactile Object Exploration with Gaussian Processes},'' in
  \emph{IEEE/RSJ Intl. Conf. on Intelligent Robots and Systems}, 2016.

\bibitem{lu-ram2020-MultiFingeredGP}
Q.~Lu, M.~V. der Merwe, B.~Sundaralingam, and T.~Hermans, ``Multi-fingered
  grasp planning via inference in deep neural networks,'' \emph{{IEEE} Robotics
  \& Automation Magazine}, 2020.

\bibitem{ciocarlie-ijrr2009-eigengrasp}
M.~T. Ciocarlie and P.~K. Allen, ``Hand posture subspaces for dexterous robotic
  grasping,'' \emph{Intl. Journal of Robotics Research}, vol.~28, no.~7, 2009.

\bibitem{suarez-2009tro-contact-regions}
M.~A. Roa and R.~Suarez, ``Computation of independent contact regions for
  grasping 3-d objects,'' \emph{{IEEE} Trans. on Robotics}, vol.~25, no.~4,
  2009.

\bibitem{suarez-2011ijrr-synthesizing-grasps}
C.~Rosales, L.~Ros, J.~M. Porta, and R.~Su{\'a}rez, ``Synthesizing grasp
  configurations with specified contact regions,'' \emph{Intl. Journal of
  Robotics Research}, vol.~30, no.~4, 2011.

\bibitem{zheng-2005ijrr-force-closure-uncertainty}
Y.~Zheng and W.-H. Qian, ``Coping with the grasping uncertainties in
  force-closure analysis,'' \emph{Intl. Journal of Robotics Research}, vol.~24,
  no.~4, 2005.

\bibitem{chen-2018-pSDF}
D.~Chen, V.~Dietrich, Z.~Liu, and G.~von Wichert, ``A probabilistic framework
  for uncertainty-aware high-accuracy precision grasping of unknown objects,''
  \emph{Journal of Intelligent \& Robotic Systems}, vol.~90, no.~1, 2018.

\bibitem{hang-tro2016-hierarchical-fingertip-space}
K.~{Hang}, M.~{Li}, J.~A. {Stork}, Y.~{Bekiroglu}, F.~T. {Pokorny},
  A.~{Billard}, and D.~{Kragic}, ``Hierarchical fingertip space: A unified
  framework for grasp planning and in-hand grasp adaptation,'' \emph{{IEEE}
  Trans. on Robotics}, vol.~32, no.~4, 2016.

\bibitem{siddiqui-frontiers2021-bayesian-exploration}
M.~S. Siddiqui, C.~Coppola, G.~Solak, and L.~Jamone, ``Grasp stability
  prediction for a dexterous robotic hand combining depth vision and haptic
  bayesian exploration,'' \emph{Frontiers in Robotics and AI}, 2021.

\bibitem{morales-iros2006-asfour}
A.~Morales, T.~Asfour, P.~Azad, S.~Knoop, and R.~Dillmann, ``Integrated grasp
  planning and visual object localization for a humanoid robot with
  five-fingered hands,'' in \emph{IEEE/RSJ Intl. Conf. on Intelligent Robots
  and Systems}, 2006.

\bibitem{platt2011simultaneous}
R.~Platt, L.~Kaelbling, T.~Lozano-Perez, and R.~Tedrake, ``{Simultaneous
  Localization and Grasping as a Belief Space Control Problem},'' in
  \emph{{Intl. Symposium on Robotics Research}}, vol.~2, 2011.

\bibitem{kappler-icra2015-leveraging-big-data}
D.~{Kappler}, J.~{Bohg}, and S.~{Schaal}, ``Leveraging big data for grasp
  planning,'' in \emph{IEEE Intl. Conf. on Robotics and Automation}, 2015.

\bibitem{varley-iros2015-generating-multi-fingered}
J.~{Varley}, J.~{Weisz}, J.~{Weiss}, and P.~{Allen}, ``Generating
  multi-fingered robotic grasps via deep learning,'' in \emph{IEEE/RSJ Intl.
  Conf. on Intelligent Robots and Systems}, 2015.

\bibitem{saxena-aaai2008-learning-grasp}
A.~Saxena, L.~L.~S. Wong, and A.~Y. Ng, ``Learning grasp strategies with
  partial shape information,'' in \emph{AAAI National Conf. on Artificial
  Intelligence}, 2008.

\bibitem{kopicki-ijrr2019-better-generative-models}
M.~S. Kopicki, D.~Belter, and J.~L. Wyatt, ``Learning better generative models
  for dexterous, single-view grasping of novel objects,'' \emph{Intl. Journal
  of Robotics Research}, vol.~38, no. 10-11, 2019.

\bibitem{merzic-icra2019-leveraging-contact-forces}
H.~{Merzić}, M.~{Bogdanović}, D.~{Kappler}, L.~{Righetti}, and J.~{Bohg},
  ``Leveraging contact forces for learning to grasp,'' in \emph{IEEE Intl.
  Conf. on Robotics and Automation}, 2019.

\bibitem{veres-ral17-grasp-motor-imagery}
M.~Veres, M.~Moussa, and G.~W. Taylor, ``Modeling grasp motor imagery through
  deep conditional generative models,'' \emph{{IEEE Robotics and Automation
  Letters}}, vol.~2, no.~2, 2017.

\bibitem{shao-ral2020-unigrasp}
L.~Shao, F.~Ferreira, M.~Jorda, V.~Nambiar, J.~Luo, E.~Solowjow, J.~A. Ojea,
  O.~Khatib, and J.~Bohg, ``Unigrasp: Learning a unified model to grasp with
  multifingered robotic hands,'' \emph{{IEEE Robotics and Automation Letters}},
  vol.~5, no.~2, 2020.

\bibitem{sundermeyer-icra2021-contact-graspnet}
M.~Sundermeyer, A.~Mousavian, R.~Triebel, and D.~Fox, ``Contact-graspnet:
  Efficient 6-dof grasp generation in cluttered scenes,'' in \emph{IEEE Intl.
  Conf. on Robotics and Automation}, 2021.

\bibitem{wu-iros2019-pixel-attentive}
B.~{Wu}, I.~{Akinola}, and P.~K. {Allen}, ``Pixel-attentive policy gradient for
  multi-fingered grasping in cluttered scenes,'' in \emph{IEEE/RSJ Intl. Conf.
  on Intelligent Robots and Systems}, 2019.

\bibitem{lundell-2021ral-clutter}
J.~Lundell, F.~Verdoja, and V.~Kyrki, ``Ddgc: Generative deep dexterous
  grasping in clutter,'' \emph{{IEEE Robotics and Automation Letters}}, vol.~6,
  no.~4, 2021.

\bibitem{mousavian-iccv2019-graspnet}
A.~Mousavian, C.~Eppner, and D.~Fox, ``6-dof graspnet: Variational grasp
  generation for object manipulation,'' in \emph{Intl. Conf. on Computer
  Vision}, 2019.

\bibitem{duan-frontiers2021-dexterous-grasping-summary}
H.~Duan, P.~Wang, Y.~Huang, G.~Xu, W.~Wei, and X.~Shen, ``Robotics dexterous
  grasping: The methods based on point cloud and deep learning,''
  \emph{Frontiers in Neurorobotics}, vol.~15, 2021.

\bibitem{calandra-ral2018-more-than-feeling}
R.~{Calandra}, A.~{Owens}, D.~{Jayaraman}, J.~{Lin}, W.~{Yuan}, J.~{Malik},
  E.~H. {Adelson}, and S.~{Levine}, ``More than a feeling: Learning to grasp
  and regrasp using vision and touch,'' \emph{{IEEE Robotics and Automation
  Letters}}, vol.~3, no.~4, 2018.

\bibitem{wu-corl2019-mat}
B.~Wu, I.~Akinola, J.~Varley, and P.~K. Allen, ``{MAT: Multi-Fingered Adaptive
  Tactile Grasping via Deep Reinforcement Learning},'' in \emph{Conference on
  Robot Learning}, 2019.

\bibitem{dang-ar2014-stable-grasping}
H.~Dang and P.~K. Allen, ``Stable grasping under pose uncertainty using tactile
  feedback,'' \emph{Autonomous Robots}, vol.~36, 2014.

\bibitem{farias-ral21-gpis-tactile-vision}
C.~de~Farias, N.~Marturi, R.~Stolkin, and Y.~Bekiroglu, ``Simultaneous tactile
  exploration and grasp refinement for unknown objects,'' \emph{IEEE Robotics
  and Automation Letters}, vol.~6, no.~2, pp. 3349--3356, 2021.

\bibitem{bfgs}
R.~H. Byrd, P.~Lu, J.~Nocedal, and C.~Zhu, ``A limited memory algorithm for
  bound constrained optimization,'' \emph{SIAM Journal on Scientific
  Computing}, vol.~16, no.~5, pp. 1190--1208, 1995.

\bibitem{mordatch-siggraph2012-cio}
I.~Mordatch, Z.~Popovi\'{c}, and E.~Todorov, ``Contact-invariant optimization
  for hand manipulation,'' in \emph{SIGGRAPH}, 2012.

\bibitem{kunz-rss12-time-optimal}
T.~Kunz and M.~Stilman, ``Time-optimal trajectory generation for path following
  with bounded acceleration and velocity,'' in \emph{Robotics: Science and
  Systems}, 2012.

\bibitem{wettels2014multimodal}
N.~Wettels, J.~A. Fishel, and G.~E. Loeb, ``Multimodal tactile sensor,'' in
  \emph{The Human Hand as an Inspiration for Robot Hand Development}.\hskip 1em
  plus 0.5em minus 0.4em\relax Springer, 2014, pp. 405--429.

\bibitem{gjk}
E.~Gilbert, D.~Johnson, and S.~Keerthi, ``A fast procedure for computing the
  distance between complex objects in three-dimensional space,'' \emph{IEEE
  Journal on Robotics and Automation}, vol.~4, no.~2, 1988.

\bibitem{mcubes}
W.~E. Lorensen and H.~E. Cline, ``Marching cubes: A high resolution 3d surface
  construction algorithm,'' \emph{SIGGRAPH Comput. Graph.}, vol.~21, no.~4, pp.
  163--–169, Aug 1987.

\bibitem{trimesh}
\BIBentryALTinterwordspacing
{Dawson-Haggerty et al.}, ``trimesh.'' [Online]. Available:
  \url{https://trimsh.org/}
\BIBentrySTDinterwordspacing

\bibitem{ycb}
B.~Calli, A.~Singh, J.~Bruce, A.~Walsman, K.~Konolige, S.~Srinivasa, P.~Abbeel,
  and A.~M. Dollar, ``Yale-cmu-berkeley dataset for robotic manipulation
  research,'' \emph{Intl. Journal of Robotics Research}, vol.~36, no.~3, pp.
  261--268, 2017.

\end{thebibliography}
}
\end{document}